\documentclass[sn-nature,a4paper,top=1in,bottom=1.2in,left=1in,right=1in,sn-basic]{sn-jnl}

\usepackage{graphicx}%
\usepackage{multirow}%
\usepackage{amsmath,amssymb,amsfonts}%
\usepackage{amsthm}%
\usepackage{mathrsfs}%
\usepackage[title]{appendix}%
\usepackage{xcolor}%
\usepackage{textcomp}%
\usepackage{manyfoot}%
\usepackage{booktabs}%
\usepackage{algorithm}%
\usepackage{algorithmicx}%
\usepackage{algpseudocode}%
\usepackage{listings}%
\usepackage{subcaption}
\usepackage{multirow}
\usepackage{siunitx}
\usepackage{wrapfig}
\usepackage{lineno}

\theoremstyle{thmstyleone}%
\theoremstyle{thmstyletwo}%

\theoremstyle{thmstylethree}%

\begin{document}

\nolinenumbers

\title[When Does Global Attention Help? A Unified Empirical Study on Atomistic Graph Learning]{When Does Global Attention Help? A Unified Empirical Study on Atomistic Graph Learning}

\author{\fnm{Arindam} \sur{Chowdhury}}\email{chowdhurya1@ornl.gov}

\author*{\fnm{Massimiliano} \sur{Lupo Pasini}}\email{lupopasinim@ornl.gov}


\affil{\orgdiv{Computational Sciences and Engineering Division (CSED)}, \orgname{Oak Ridge National Laboratory}, \orgaddress{\street{5700, 1 Bethel Valley Rd}, \city{Oak Ridge}, \postcode{37830}, \state{TN}, \country{USA}}}


 
\abstract{Graph neural networks (GNNs) are widely used as surrogates for costly experiments and first-principles simulations to study the behavior of compounds at atomistic scale, and their architectural complexity is constantly increasing to enable the modeling of complex physics. While most recent GNNs combine more traditional message passing neural networks (MPNNs) layers to model short-range interactions with more advanced graph transformers (GTs) with global attention mechanisms to model long-range interactions, it is still unclear when global attention mechanisms provide real benefits over well-tuned MPNN layers due to inconsistent implementations, features, or hyperparameter tuning. We introduce the first unified, reproducible benchmarking framework—built on HydraGNN—that enables seamless switching among four controlled model classes: MPNN, MPNN with chemistry/topology encoders, GPS-style hybrids of MPNN with global attention, and fully fused local–global models with encoders. Using seven diverse open-source datasets for benchmarking across regression and classification tasks, we systematically isolate the contributions of message passing, global attention, and encoder-based feature augmentation. Our study shows that encoder-augmented MPNNs form a robust baseline, while fused local–global models yield the clearest benefits for properties governed by long-range interaction effects. We further quantify the accuracy–compute trade-offs of attention, reporting its overhead in memory. Together, these results establish the first controlled evaluation of global attention in atomistic graph learning and provide a reproducible testbed for future model development.

\paragraph*{Scientific Contribution.}
This work provides the first unified and reproducible empirical framework to systematically evaluate when global attention mechanisms yield measurable benefits over well-tuned message passing neural networks for atomistic graph learning. By implementing all combinations of message passing, encoder-based feature augmentation, and global attention within a single HydraGNN pipeline and under identical training and hyperparameter optimization protocols, we eliminate confounding effects present in prior comparisons. The results transform widely stated but weakly verified assumptions about long-range modeling in molecular GNNs into empirically testable conclusions, clarifying when global attention is beneficial and when expressive MPNNs remain sufficient.

}




\maketitle

{\footnotesize{This manuscript has been authored in part by UT-Battelle, LLC, under contract DE-AC05-00OR22725 with the US Department of Energy (DOE). The US government retains and the publisher, by accepting the article for publication, acknowledges that the US government retains a nonexclusive, paid-up, irrevocable, worldwide license to publish or reproduce the published form of this manuscript, or allow others to do so, for US government purposes. DOE will provide public access to these results of federally sponsored research in accordance with the DOE Public Access Plan (\url{http://energy.gov/downloads/doe-public-access-plan}).}}

{\paragraph{Abbreviations.}
MPNN: Message Passing Neural Network;
GT: Graph Transformer;
GPS: General, Powerful, Scalable hybrid GNN;
CE: Chemical Encoders;
TE: Topological Encoders;
LPE: Laplacian Positional Encodings.}

\section{Introduction}\label{sec:intro}

Predicting atomistic and material properties from atomistic structure is a central challenge across computational chemistry, catalysis, and materials discovery~\citep{Butler2018,Ward2016}. Traditionally, property estimation relied on (i) empirical and semi-empirical correlations—group additivity and Hammett–Taft substituent analysis, followed by QSAR/QSPR with hand-crafted descriptors~\citep{Benson1976,Hammett1937,Taft1952,Cherkasov2014}—(ii) physics-based electronic-structure methods such as Hartree-Fock (HF) density functional theory (DFT), post-HF methods such as second-order Møller-Plesset (MP2) perturbation theory and coupled cluster singles, doubles, (triples) CCSD(T))~\citep{Kohn1965,ParrYang1989,Mardirossian2017,Bartlett2007,Grimme2010}, (iii) molecular dynamics with empirical or \emph{ab initio} potentials~\citep{FrenkelSmit2002}, and (iv) direct experiments and high-throughput screening~\citep{Gregoire2013}. While foundational, these approaches face limitations: QSAR and group-additivity depend on fixed descriptors with narrow domains of applicability and often miss nonlocal, conformational, or environment effects; DFT/post-HF deliver systematic accuracy but scale steeply (roughly $\mathcal{O}(N^{3\text{--}4})$ for Kohn–Sham DFT and $\mathcal{O}(N^7)$ for CCSD(T)) and can struggle with dispersion or strong correlation without specialized corrections~\citep{Mardirossian2017,Grimme2010,Bartlett2007}; MD accuracy hinges on force-field fidelity and long trajectories to sample rare events \citep{FrenkelSmit2002}; and experiments are costly, lower-throughput, and condition-specific \citep{Gregoire2013}. In contrast, modern machine learning (ML)-based methods amortize expensive computation/measurement into training and then yields fast, scalable inference while capturing complex, nonlocal structure–property relationships from 2D/3D atomistic representations~\citep{Behler2007,Rupp2012,Smith2017,Butler2018,Ward2016}.

Graph-based learning offers a natural paradigm: a molecule can be modeled as a graph whose nodes (atoms) and edges (bonds or distance-thresholded connections) carry domain-specific attributes, optionally augmented with 3D coordinates in geometric graphs \citep{Duvenaud2015,Kearnes2016,Gilmer2017MPNN,Schutt2018,Klicpera2020}. Such a structure supports inductive biases -- permutation invariance and, when needed, euclidean equivariance -- that are essential for modeling scalar, vector, and tensor targets \citep{Zaheer2017,Bronstein2021,Thomas2018,Fuchs2020SE3Transformer,Satorras2021}. Graph Neural Networks (GNN)~\citep{Battaglia2018GraphNetworks,Zhou2020GNNReview,Wu2021GNNSurvey} including both message-passing neural networks (MPNN)~\citep{Gilmer2017MPNN} and their geometric extensions (e.g., SchNet, DimeNet, EGNN)~\citep{Schutt2018,Klicpera2020DimeNet,Satorras2021} can then leverage the connectivity structure for predicting atomistic properties.

Despite their success, MPNNs can struggle when target properties depend on \emph{long-range interactions} (LRI)—such as electrostatics, induction, and dispersion—that persist beyond small $k$-hop neighborhoods or bond-local connectivity \citep{London1937,CasimirPolder1948,Israelachvili2011ISF3e,Stone2013TIMF}.
Particularly, nonlocal effects due to LRI arise when influential atoms are far in either sense and recaps their physical scalings (e.g., $-C_6/R^6$ dispersion) \citep{London1937,Stone2013TIMF}. Deepening MPNNs to enlarge the receptive field invites two well-known pathologies, \emph{over-smoothing} and \emph{over-squashing}~\citep{Li2018,OonoSuzuki2020,AlonYahav2021,Topping2022}. These limitations argue for mechanisms that propagate information globally without sacrificing local structural details. Transformer-style global attention provides such a mechanism \citep{vaswani2017attention}. Graph Transformers (GTs) replace fixed-radius aggregation with learnable all-to-all information flow via multi-head self-attention, potentially capturing both local and global dependencies in a single layer~\citep{dwivedi2020generalization,kreuzer2021rethinking,Ying2021Graphormer}. However, naively applying dense attention over $N$ nodes incurs $\mathcal{O}(N^2)$ time and memory per layer, which can be prohibitive for large atomistic graphs~\citep{vaswani2017attention}. Hybrid designs (e.g., GPS~\citep{rampasek2022gps} have been proposed to combine the modeling capabilities of strong local message passing with global attention mechanisms that sparsify the connectivity of the graph to contain their computational cost. However, these methods have only been evaluated on a small group of datasets and over a limited class of MPNNs as local information aggregators.

In this work, we present a unified, reproducible framework built on top of HydraGNN~\citep{lupopasini2022hydragnn} -- a highly parallelized graph multitask learning pipeline -- for systematically evaluating transformer-based graph models for atomistic property prediction alongside a large class of SOTA MPNNs, including equivariant MPNNs. The framework instantiates four controlled configurations: (i) MPNN, (ii) MPNN with chemistry-/topology-based encoders, (iii) GPS-style MPNN with global attention, and (iv) a fused local–global model with domain-specific encoders. {In this work, ``controlled'' does not mean fixing hyperparameters across architectures, which would bias results toward suboptimal configurations. Instead, control is enforced at the level of (i) identical datasets and splits, (ii) identical training pipelines and budgets, (iii) identical HPO search spaces conditioned on architectural constraints, and (iv) identical model selection criteria based on validation performance. This ensures that performance differences reflect architectural capabilities rather than tuning artifacts.}
The compact encoder suite supplies chemically and structurally informative inputs: per-atom physico-chemical descriptors, node/edge structural features (e.g., degree, centralities, clustering, $k$-core, edge indices), and Laplacian positional encodings; edges additionally carry bond and, when available, distance information. Lightweight linear embeddings integrate these channels, and training uses distributed data parallelism (DDP) with automated hyperparameter optimization to enable fair ablations. 
Across twelve benchmarks spanning regression and classification, we assess (1) when global attention yields gains beyond well-tuned MPNNs, (2) the contribution of domain/topological encoders, and (3) the accuracy–compute trade-off associated with the quadratic cost of attention. Empirically, encoders systematically improve performance while attention alone is not uniformly superior. Fused local–global models provide the largest benefits on tasks with significant nonlocal or geometric effects, while encoder-augmented MPNNs remain competitive on others. The result is a standardized experimental setup and sets the ground towards clarifying when opting for global attention is necessary, thereby serving as a reference point for future method development.

\subsection{Related Work}\label{sec:related}

{Attention mechanisms have been previously incorporated directly into message passing neural networks to modulate the aggregation of neighboring messages. Notable examples include Graph Attention Networks (GAT) \cite{Velickovic2018GAT}, which replace uniform neighborhood aggregation with learned attention weights over edges, enabling adaptive importance weighting of neighbors during message passing. Subsequent works have extended this idea by integrating attention into more general MPNN formulations, including edge-conditioned or directional attention, attention over message channels, and attention mechanisms coupled with learned edge representations \cite{10.1145/3534678.3539296, brody2022how}. These approaches primarily employ local attention, where attention weights are computed over immediate neighbors and integrated within the message passing operator itself.}

GTs for molecules have seen several recent advancements. \emph{Graphormer} injects structural biases (shortest-path, centrality, edge encodings) and achieves strong OGB results, but relies on dense quadratic attention and task-specific heuristics without controlled comparisons to equally tuned MPNNs~\citep{Ying2021}. \emph{MAT} augments attention with RDKit features and pairwise distance biases, yet similarly inherits the quadratic cost and heavy feature engineering~\citep{Maziarka2020}. Equivariant Transformers (e.g., \emph{EquiformerV2}) improve 3D fidelity for atomistic energies/forces but are computationally demanding and targeted to continuous 3D labels rather than the broader mix of graph\mbox{-}level and node/edge targets considered here~\citep{EquiformerV2}. Large-scale frameworks such as \emph{Uni-Mol}~\citep{zhou2023unimol} leverage massive SE(3)-equivariant pretraining to obtain strong downstream accuracy, but conflate architectural gains with data scale and are difficult to reproduce under typical compute budgets \citep{UniMol}. Hybrid local–global designs provide a general recipe that couples message passing with global attention, though evaluations emphasize non-atomistic benchmarks and do not isolate chemistry-aware encoders~\citep{Rampasek2022}; follow-up work (\emph{GPS++}) on PCQM4Mv2 shows that well-tuned message passing can retain much of the benefit with little or no attention, but focuses on a single large dataset \citep{Masters2023}. Finally, long-range benchmarks (LRGB) motivate global receptive fields yet are not atomistic-specific and omit common chemical settings such as node\mbox{-}level charges or multi-label assays \citep{LRGB}. Consequently, the field lacks a controlled, reproducible study that compares these architectural choices under a common framework.

{Multitask learning has been widely explored in molecular machine learning as a means to improve data efficiency, generalization, and representation sharing across related chemical properties \cite{ramsundar2015massivelymultitasknetworksdrug, martin2025massivelymultitask}. Early studies showed that jointly learning multiple molecular endpoints using shared graph-based representations can outperform single-task models, particularly when tasks are correlated or when individual datasets are small \cite{capela2019multitasklearninggraphneural}. Subsequent work has extended multitask learning to graph neural networks and message passing architectures \cite{10.1145/3580305.3599265, chen2025multitaskgnn}, showing that shared encoders combined with task-specific prediction heads can effectively capture common chemical structure–property relationships while preserving task-specific expressivity. However, prior studies often differ substantially in model architecture, training protocols, and dataset choices, making it difficult to disentangle the effects of multitask learning from other confounding factors.}

\subsection{Contributions}\label{sec:contri}
{In contrast to state-of-the-art methods, our work does not propose a new attention mechanism for MPNNs or graph-based transformer architecture. Instead, we provide a controlled and systematic evaluation of multiple attention integration strategies, ranging from local message passing, encoder-based feature augmentation, and global attention—within a unified HydraGNN framework under identical training and hyperparameter optimization protocols. Moreover, we combine the unified framework of HydraGNN for global attention with multitask learning strategies and evaluate the integration of the two techniques. }

Unlike prior studies that test isolated architectures on narrow benchmarks, we systematically evaluate four model classes under a single experimental pipeline, allowing clear attribution of accuracy gains to either architecture, encoders, or their combination. To that end, we make the following key contributions:
\begin{itemize}
\item \textbf{Development of a unified framework to combine MPNN and GT.} We expanded the existing HydraGNN-based pipeline that provides MPNN architectures with GT architectures, and offer seamless selection of model pathways viz. MPNN-only, MPNN\,+\,Encoders, GPS-style, and fully fused (MPNN\,+\,GPS\,+\,Encoders) configurations with DDP and automated HPO (Fig.~\ref{fig:block}).
\item \textbf{Inclusion of Domain/topology encoders into graph embeddings.} Chemically informed per-atom descriptors, node/edge structural encodings, and Laplacian positional encodings, integrated via lightweight embeddings (Sections~\ref{sec:encoders}-\ref{sec:feature-embedder}). 
\item \textbf{Comprehensive evaluation with ablation studies.} Cross-domain study on $7$ datasets (Table~\ref{tab:datasets}) with regression, multi-class, and multi-label tasks; quantitative comparisons (Sections~\ref{sec:zinc}-\ref{sec:pcba}). 
\end{itemize}

\subsection{Paper Organization}
Section~\ref{sec:background} reviews preliminaries on graphs, geometric learning, and invariance/equivariance, and motivates global attention through LRIs and MPNN limitations, then introduces GPS-style hybrids. Section~\ref{sec:method} describes the proposed framework including encoder and embedding modules. Section~\ref{sec:exp} reports experiments, with Section~\ref{sec:disc} discussing implications and Section~\ref{sec:conc} concluding the work.

\section{Background and Problem Setting}\label{sec:background}

We first introduce some fundamental definitions and concepts that will be extensively used throughout this work. $\mathcal{X}$ represents sets. $[\mathbf{X}]_{uv}$, and $[\mathbf{x}]_{i}$ denote the entries of a multi-dimensional matrix $\mathbf{X}$, and a vector $\mathbf{x}$. The generic subindex $:$ denotes a whole dimension, e.g., row $i$ of matrix $\mathbf{X}$ is denoted as $[\mathbf{X}]_{i \,:}$. 
Scalars are denoted by $x$ or $X$.

\subsection{Atomistic Graphs} A graph is a data structure that represents entities (nodes) and their relations (edges). Formally, $\mathcal{G} = \{\mathcal{V},\mathcal{E}\}$ denotes an undirected graph where $\mathcal{V}$ represents the set of nodes and $\mathcal{E}$ represents the set of edges. $N = |\mathcal{V}|$ is the number of nodes in the graph. Each node $v \in \mathcal{V}$ supports a $p-$dimensional feature vector denoted by $\mathbf{x}_v \in \mathbb{R}^p$, such that $[\mathbf{X}]_{v\,:} = \mathbf{x}_v$, while $\mathbf{e}_{uv} \in \mathbb{R}^f$ represents edge attributes supported by each edge $(u,v) \in \mathcal{E}$ where $[\mathbf{E}]_{uv\,:} = \mathbf{e}_{uv}$. Additionally, the graph $\mathcal{G}$ can be defined as a \textit{geometric graph} if each node $v$ represents a point on the 3D Euclidean space and their connections have a notion of distance associated with them. 
\begin{wrapfigure}{r}{0.3\columnwidth} 
    \centering
    \vspace{-15pt} 
    \resizebox{\linewidth}{!}{%
   \includegraphics[width=0.2\linewidth]{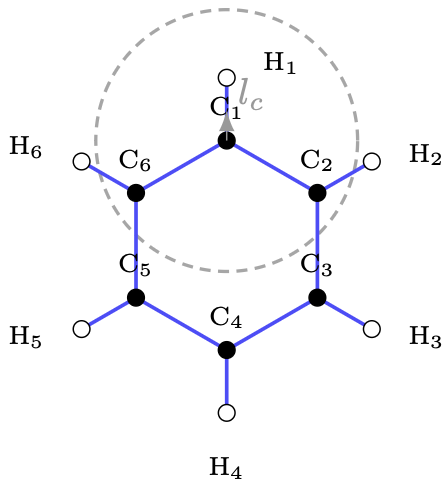}}
  \caption{Geometric atomistic graph with cutoff $r_c$. }
  \label{fig:mor_cutoff}
  \vspace{-15pt}
\end{wrapfigure}
In this case, the augmented graph is given by $\mathcal{G} = \{\mathcal{V},\mathcal{E},\mathbf{X},\mathbf{E},\mathbf{R}\}$ where $\mathbf{R} \in \mathbb{R}^{N\times 3}$ stores the 3D coordinates of the nodes. Often, $\mathcal{E}$ is built by thresholding the distance between the neighboring nodes using a radius cutoff ($r_c$) specific to a given downstream application. Particularly,
$$
\mathcal{E} = \{ (u,v) \mid u \neq v, \ \| \mathbf{r}_u - \mathbf{r}_v \| < r_c \}
$$
where $[\mathbf{R}]_{u:} = \mathbf{r}_u$. Ex., for atomistic graphs such as Fig.~\ref{fig:mor_cutoff}, the radius cutoff is carefully chosen to capture chemical and physical properties including specific bond lengths ($1–2\,\text{\AA}$), non-bonded interactions like van der Waals forces ($3–5\,\text{\AA}$) and metal-ligand interactions ($5–6\,\text{\AA}$).
Leveraging the connectivity structure of graphs to generate useful mathematical abstractions has been the fundamental motivation in developing the theory of graph machine learning and specifically GNNs.

\subsection{Long Range Interactions}\label{sec:lri}

The long-range interaction (LRI) effects in molecules arise from fundamental physical forces—electrostatics (multipole interactions), induction (polarization), and dispersion (London forces)—which decay only as inverse powers of the interatomic distance and therefore persist well beyond the typical neighbor cutoffs used in local models \citep{Stone2013TIMF,Israelachvili2011ISF3e,London1937GeneralTheory}. To formalize LRIs, we consider two complementary notions of distance between atoms $u$ and $v$: the shortest-path (graph) distance $d_G(u,v)$ on the atomistic bond graph, and the Euclidean distance $r_{uv} = \lVert \mathbf{r}_u - \mathbf{r}_v \rVert_2$ in 3D space \citep{Trinajstic1992ChemicalGraphTheory}. An interaction is deemed \emph{long-range} for a given node $u$ if there exists at least one atom $v$ whose contribution to the target property cannot be captured by information within a small $k$-hop neighborhood:
\[
\exists v:\ d_G(u,v) \gg k \quad\text{or}\quad r_{uv} \gg r_c \quad\text{but}\quad \text{influence}(v\!\to\!u) \not\approx 0.
\]
These two distances capture different manifestations of nonlocality: in folded or spatially compact conformers, $d_L$ can be small even when $d_G$ is large, producing through-space, nonbonded contacts; in extended or rigid systems, $d_G$ may be small while $d_L$ is large. Related concepts appear in protein \emph{contact order}, which measures spatial proximity despite large sequence (graph) separation \citep{Plaxco1998ContactOrder}.  

Physically, such atom--atom relationships can be grouped into interactions between two molecular \emph{fragments} $A$ and $B$ containing $u$ and $v$, respectively. At large separations, the interaction energy between $A$ and $B$ admits a multi-pole--polarization--dispersion decomposition:
\[
E_{\text{int}}(R) \;\approx\; E_{\text{elst}}(R) \;+\; E_{\text{ind}}(R) \;+\; E_{\text{disp}}(R) \;+\; \cdots ,
\]
where $R$ is the Euclidean separation between the fragments’ centers (or a representative interatomic distance). Electrostatics is described by a multi-pole expansion, where the potential from a monopole decays as $R^{-1}$, from a dipole as $R^{-2}$, and in general from a multi-pole of order $\ell$ as $R^{-(\ell+1)}$, with the corresponding fields decaying one power faster (e.g., a dipole field as $R^{-3}$) \citep{Stone2013TIMF}. Induction arises when the field of one fragment polarizes the other; for isotropic fragments, the orientationally averaged induction energy scales as $R^{-6}$ \citep{Stone2013TIMF,Israelachvili2011ISF3e}. Dispersion, due to correlated electron fluctuations, has a leading attractive term $-C_6/R^6$ with higher-order corrections $-C_8/R^8$, $-C_{10}/R^{10}$ \citep{London1937GeneralTheory,Stone2013TIMF}, and in the retarded Casimir--Polder regime scales as $R^{-7}$ \citep{CasimirPolder1948Retardation}. By contrast, short-range exchange (Pauli) repulsion decays exponentially and is described in frameworks such as SAPT and in overlap-based models \citep{Jeziorski1994SAPT,BornMayer1932ZPhys}.  


\subsection{MPNN Architectures} An MPNN architecture~\citep{Battaglia2018GraphNetworks,Zhou2020GNNReview,Wu2021GNNSurvey} is a deep learning architecture that performs convolution-like operations on node features by exploiting the irregular connectivity patterns inherent in graph-structured data. A $K-$layer MPNN is given by,
\begin{align}
    \bar{\mathbf{H}}^{(k)} &= \bigoplus(\phi(\mathbf{H}^{(k-1)}, \omega_k), \mathcal{E}, \mathbf{E}) \,\,\forall k \in \{1,\dots,K\}\\
    \mathbf{H}^{(k)} &= \Psi(\bar{\mathbf{H}}^{(k)},\mathbf{H}^{(k-1)})
\end{align}
where $\mathbf{H}^{(k)} \in \mathbb{R}^{N\times d_{k}}$ are the node representations after $k$ MPNN layers and $\mathbf{H}^{(0)} = \mathbf{X}$. $\phi(\cdot,\omega)$ denotes a linear transformation of the node features and the trainable weights $\omega_k$ are shared by all nodes in the graph at a given layer $k$. 
\begin{wrapfigure}{r}{0.6\columnwidth} 
    \centering
    \vspace{-10pt} 
    \resizebox{\linewidth}{!}{%
   \includegraphics[width=0.7\linewidth]{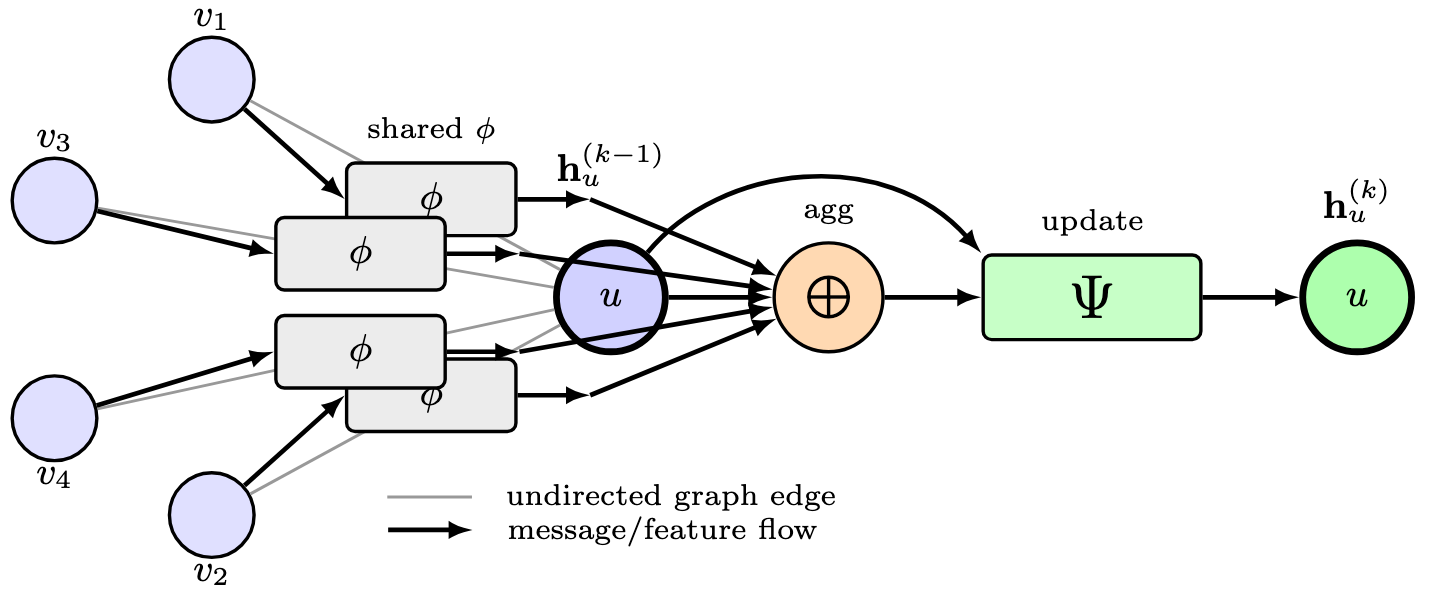}}
  \caption{A generic MPNN layer $k$ is detailed. $\mathbf{h}_u = [\mathbf{H}]_u$ }
  \label{fig:gnn_layer}
  \vspace{-10pt}
\end{wrapfigure}
The aggregator $\bigoplus$ generates node representations by combining information from corresponding local neighborhoods $\mathcal{N}_u = \{v \mid v\in \mathcal{V}, v \neq u, (u,v) \in \mathcal{E}\}$, while leveraging the edge attributes $\mathbf{E}$, if available. An arbitrary MPNN layer is shown in Fig.~\ref{fig:gnn_layer}.
Depending on the specific functional form of $\bigoplus$, a broad class of MPNN models~\citep{KipfWelling2017GCN,Hamilton2017GraphSAGE,Gilmer2017MPNN,Velickovic2018GAT,Xu2019GIN,Battaglia2018GraphNetworks,Wu2021GNNSurvey,Zhou2020GNNReview} have been proposed. $\Psi$ is a node update function such as a multi-layer perceptron (MLP) or gated recurrent unit (GRU).
The MPNN output $\mathbf{H}^{(K)}$ stores the full set of node representations that are used for node\mbox{-}level downstream tasks. For graph\mbox{-}level tasks, several pooling operations (i.e., min, max, sum, average) have been employed to convert $\mathbf{H}^{(K)}$ into a dense graph\mbox{-}level representation.

In atomistic modeling applications, it is crucial to incorporate geometric information such as atomic coordinates, interatomic distances, and angles into the message-passing process. To this aim, chemistry or physics-driven MPNN architectures like SchNet~\citep{Schutt2018SchNet}, DimeNet~\citep{Klicpera2020DimeNet}, and EGNN~\citep{Satorras2021EGNN} extend standard MPNNs by encoding continuous geometric features as edge attributes, applying distance-based filters, or enforcing equivariance under Euclidean transformations. These chemistry-inspired MPNN architectures construct geometric embeddings by:
\begin{itemize}
\item {Extracting pairwise distances and triplet bond angles for nodes} $u$ {and} $v$ {with coordinates} $\mathbf{r}_u,\mathbf{r}_v \in \mathbb{R}^3$, 
\begin{gather}
r_{uv} = \|\mathbf{r}_u - \mathbf{r}_v\|, 
\quad 
\theta_{uvw} = \angle(\mathbf{r}_v-\mathbf{r}_u,\ \mathbf{r}_w-\mathbf{r}_u), \label{eq:geo_defs}
\end{gather}
\item {Embedding these geometric quantities into a latent edge representation}:
\begin{gather}
\bar{\mathbf{e}}_{uv} = \phi_l(r_{uv}) \;+\; \sum_{w\in \mathcal{N}_u\setminus\{v\}} \phi_a(\theta_{uvw}), \label{eq:edge_embed}
\end{gather}
{where} $\phi_r$ and $\phi_a$ {are learnable radial and angular embedding functions, respectively}.\\
\item {Computing messages from neighbor} $v$ {to node} $u$ at any layer $k$ of the $K-$layer model as:
\begin{gather}
\mathbf{m}^{(k)}_{u\leftarrow v} = \xi\!\big(\mathbf{h}^{(k-1)}_u, \mathbf{h}^{(k-1)}_v,\ \bar{\mathbf{e}}_{uv}\big), \label{eq:message}
\end{gather}
{where} $\xi$ {is a trainable message function that conditions on both node features and geometry-aware edge features} and $[\mathbf{H}]_v = \mathbf{h}_v$.\\
\item Aggregating node features using incoming messages:
\begin{gather}
\mathbf{h}^{(k)}_u = \Psi\!\Big(\mathbf{h}^{(k-1)}_u,\ \bigoplus_{v\in \mathcal{N}_(u)} \mathbf{m}^{(k)}_{u\leftarrow v}\Big), \label{eq:update}
\end{gather}
\end{itemize}
For graph\mbox{-}level prediction tasks, the final node states $\mathbf{H}^{(K)}$ are aggregated via pooling layers. These approaches allow the network to remain sensitive to spatial relationships while respecting physical symmetries, thereby improving its ability to learn from data where geometry directly governs the target properties. 
While MPNNs are sufficiently expressive to  modeling complex, inter-connected data and offer better generalization capabilities than non-graph ML models (with exceptions; see Section~\ref{sec:background}), critical challenges still remain unaddressed in terms of large-scale data handling, model scalability and systematic comparison between different graph-based DL architectures.

\subsection{Invariance and Equivariance in MPNN Architectures} 
In geometric GNNs, \emph{invariance} means that the GNN prediction is unchanged under translations, rotations, or reflections of the input, whereas \emph{equivariance} means that the prediction transforms in the same way as the input under a symmetry group such as $E(n)$ or $\mathrm{SE}(3)$.
For rigid motions $(R,\mathbf{t})\in\mathrm{SE}(3)$ acting on $\mathbf{R}$ row‑wise,
invariance means $f(R\mathbf{R}+\mathbf{t})=f(\mathbf{R})$,
equivariance means $g(R\mathbf{R}+\mathbf{t})=R\,g(\mathbf{R})$.
Generally, $E(n)$/\,$\mathrm{SE}(3)$--equivariance is required when the target is directional or tensorial—e.g., atomic forces, dipole moments, and vector fields—where architectures like EGNN, SE(3)-Transformer, PaiNN, and NequIP are designed to respect these symmetries \citep{Satorras2021EGNN,Fuchs2020SE3Transformer,Schutt2021,Batzner2022}.
For invariant scalar properties (e.g., total energy, band gap, class labels), enforcing invariance and injecting \emph{invariant geometric biases}—pairwise distances and angular/triplet features—typically suffices; SchNet uses continuous filters over interatomic distances, while DimeNet/DimeNet++ augment with angles via spherical Bessel and spherical harmonic bases \citep{Schutt2018SchNet,Klicpera2020a,Klicpera2020b}.
Note that even invariant energy models can yield \emph{equivariant} forces by taking analytic gradients of the predicted energy with respect to the coordinates of the atoms' nuclei, though explicitly equivariant models often improve accuracy and sample efficiency for a more general class of vector/tensor target properties \citep{Schutt2018SchNet,Batzner2022,Satorras2021EGNN}.
\subsection{Limitations of Message-Passing}\label{sec:localMPNN}
A straightforward technique to capture LRI in MPNNs is to deepen the DL architecture to ensure that the effective receptive field covers the whole graph around the node being processed. This approach, however, entails some major shortcomings:\\

\noindent \textbf{Over-smoothing.} Stacking many neighborhood-averaging layers repeatedly applies a (normalized) smoothing operator on the graph. As depth grows, node features from the same connected component can become nearly indistinguishable, hurting discriminability of the model. Essentially, the \textit{low-pass} nature of the aggregation operation filters out most of the high-frequency discriminative information with increase in depth of the model. This phenomenon—\emph{over-smoothing}—has been analyzed both empirically via Laplacian smoothing interpretations and theoretically via asymptotic collapse of node embeddings~\citep{Li2018DeeperInsights, Oono2020GNNExpressive}.\\

\noindent \textbf{Over-squashing (bottlenecks and curvature).} Even when $K$ is large enough to cover long graph distances, information arriving from exponentially many $k$-hop nodes may be \emph{squashed} into fixed-width messages that must traverse narrow graph cuts (small edge separators). This \emph{over-squashing} bottleneck limits long-range dependency modeling in MPNNs~\citep{Alon2020Bottleneck}. Recent theory connects over-squashing to discrete curvature: negatively curved (tree-like) regions expand neighborhoods exponentially while admitting small cuts, exacerbating information congestion.\\

\noindent Therefore, local message-passing is often inadequate in capturing LRI, especially for large-sized atomistic graphs that require a larger receptive field to cover the whole graph around each node.

\subsection{GT Architectures}
In recent years, graph-based DL has successfully ventured beyond MPNN models -- which are inherently localized information aggregators and treat a graph as the combination of localized parts -- and leveraged DL models that have a global scope and can process the graph as a whole. 
Specifically, a GT~\citep{dwivedi2020generalization,kreuzer2021rethinking,ying2021transformers} is a DL architecture that uses the \textit{self-attention} mechanism~\citep{vaswani2017attention}. Note that the term ``Graph Transformer" can also refer to a specific global attention architecture~\citep{dwivedi2020generalization}. However, in this work we use it as a descriptor for the general attention-based model class. In this framework, each node aggregates information from all other nodes in the graph via a learned attention mechanism, enabling the learning of both local and global dependencies beyond fixed-hop neighborhoods. The \emph{multi-head attention} (MHA) mechanism allows the model to jointly attend to information from different representation subspaces, thereby enhancing expressiveness and robustness. 
At an arbitrary layer $m$ of an $M-$layered GT, the node embedding matrix $\mathbf{H}^{m-1} \in \mathbb{R}^{N \times d_{m-1}}$ is projected by the attention heads into query, key, and value spaces:
\begin{equation}
    \mathbf{Q}^m = \mathbf{H}^{m-1} \mathbf{W}_Q^m, \quad
    \mathbf{K}^m = \mathbf{H}^{m-1} \mathbf{W}_K^m, \quad
    \mathbf{V}^m = \mathbf{H}^{m-1} \mathbf{W}_V^m,
\end{equation}
where $\mathbf{W}_Q^m, \mathbf{W}_K^m, \mathbf{W}_V^m \in \mathbb{R}^{d_{m-1} \times d_{m}}$ are learnable weight matrices. The scaled dot-product attention for one head is computed as:
\begin{equation}
    \mathrm{AttHead}(\mathbf{Q}^m, \mathbf{K}^m, \mathbf{V}^m) = \mathrm{softmax} \left( \frac{\mathbf{Q}^m {\mathbf{K}^m}^\top}{\sqrt{d_{m}}} \right) \mathbf{V}^m.
\end{equation}
where softmax is applied independently on each row. For $B$ heads, the outputs are concatenated and linearly projected:
\begin{equation}
    \mathrm{MultiHead}(\mathbf{Q}^m, \mathbf{K}^m, \mathbf{V}^m) = 
    \mathrm{Concat}(\mathrm{AttHead}_1, \dots, \mathrm{AttHead}_B) \mathbf{W}_O^m,
\end{equation}
where $\mathbf{W}_O^m \in \mathbb{R}^{Bd_{m} \times d_m}$ is a learnable projection matrix. 
By integrating \textit{structural} and \textit{positional encodings}, the GT effectively incorporates relational inductive biases, making it applicable to diverse tasks such as node classification, link prediction, and graph\mbox{-}level prediction. We will discuss this in more detail in Section~\ref{sec:background}.

\subsection{HydraGNN}
HydraGNN~\citep{lupopasini2022hydragnn} is a multitask learning MPNN architecture designed to simultaneously predict both global (graph\mbox{-}level) and atomic (node\mbox{-}level) material properties from atomic structure inputs. 
HydraGNN efficiently integrates DDP for large-scale atomistic datasets (sizes can range from a few hundred thousand to a few millions) with a broad set of MPNN layers as base learners that can be distributed across distributed computing resources with model parallelism. 
HydraGNN employs shared MPNN layers to extract features common across all target properties, followed by multiple task-specific output decoding heads for property-specific learning via hard parameter sharing of the MPNN layers. Each task is associated with its own loss function, and the global objective is defined as a weighted sum of individual task losses.
HydraGNN supports seamless switching across SOTA message-passing, geometric and {equivariant} GNN architectures, thereby allowing for efficient architecture search through integrated HPO.
The effectiveness of this framework both in terms of prediction accuracy and scalability has been well established through rigorous evaluation on multiple material science applications~\cite{lupo_gcnn, lupopasini2022hydragnn, choi2022scalable, baker2023invariant, LUPOPASINI2023112141, LupoPasini2025HydraGNN, LUPOPASINI2025113908}.




\subsection{Hybrid GNN Architectures}\label{sec:globalatt}

General, Powerful, Scalable (GPS) framework introduces a hybrid graph architecture that combines the strengths of MPNNs and GTs. At each layer, GPS\footnote{\textbf{Complexity.}
Full self-attention over $N$ nodes (dense $N{\times}N$ attention) costs
$\mathcal{O}(N^2)$ time and memory per layer; message passing scales with
$\mathcal{O}(|\mathcal{E}|)$ (often $\mathcal{O}(N)$ on bounded-degree/radius graphs).
GPS blends both, yielding $\mathcal{O}(|\mathcal{E}|+N^2)$ per layer.} integrates local neighborhood aggregation (via MPNNs, which capture edge-level and structural information) with global multi-head attention (which enables long-range dependencies) -- by leveraging positional and structural encodings, -- to overcome the expressivity limits of standard GNNs~\citep{rampasek2022gps}. Effectively, this method enables aggregation of global information in addition to local messages at each node in each layer of the architecture. Thus, it obviates the necessity (and corresponding pitfalls) of deeper GNN architectures. Formally, the $\ell^{th}$ GPS layer is given by:  
\begin{align}
\mathbf{X}^{\ell+1},\,\mathbf{E}^{\ell+1} &= \mathrm{GPS}^{\ell}(\mathbf{X}^{\ell},\mathbf{E}^{\ell},\mathcal{E})
\end{align}
where,
\begin{align}
\mathbf{\hat{X}}^{\ell+1}_{\!M},\,\mathbf{E}^{\ell+1} &= \mathrm{GNN}^{\ell}(\mathbf{X}^{\ell},\mathbf{E}^{\ell},\mathcal{E}), \\
\mathbf{\hat{X}}^{\ell+1}_{\!T} &= \mathrm{MHA}^{\ell}(\mathbf{X}^{\ell}), \\
\mathbf{X}^{\ell+1} &= \mathrm{MLP}^{\ell}\!\big(\mathbf{\hat{X}}^{\ell+1}_{\!M} + \mathbf{\hat{X}}^{\ell+1}_{\!T}\big).
\end{align}

Building on these capabilities, GPS++~\citep{masters2023gpspp} refines the balance between local and global processing for atomistic property prediction. GPS++ emphasizes a strong, well-tuned message passing module, complemented by biased self-attention informed by structural and geometric priors, such as shortest-path and 3D distance embeddings. Through extensive ablations, GPS++ shows that much of its performance can be retained even without global self-attention, highlighting that expressive message passing remains highly competitive, particularly when 3D positional data is absent.\\

\noindent In spite of the synergistic combination of local and global information, the GPS and its enhancements have been successfully employed only on a limited class of scientific materials~\citep{rampasek2022gps,masters2023gpspp} and in smaller quantity. Moreover, the baseline results were obtained using a limited set of MPNN architectures and no equivariant models were compared. In addition, it was shown that with proper hyperparameter tuning, MPNN models could still outperform  GT models and GPS~\citep{tonshoff2023did}. Therefore, the possibility of building an end-to-end framework for handling large-quantities of scientific materials from a varied set of application areas, using a broad class of MPNN models in tandem with global attention -- with best models selected using automated hyperparameter optimization (HPO) -- remains an open problem for the research community.

\section{Method}\label{sec:method}

\begin{figure}[!ht]
    \centering
    \centering
    \includegraphics[width=0.95\linewidth]{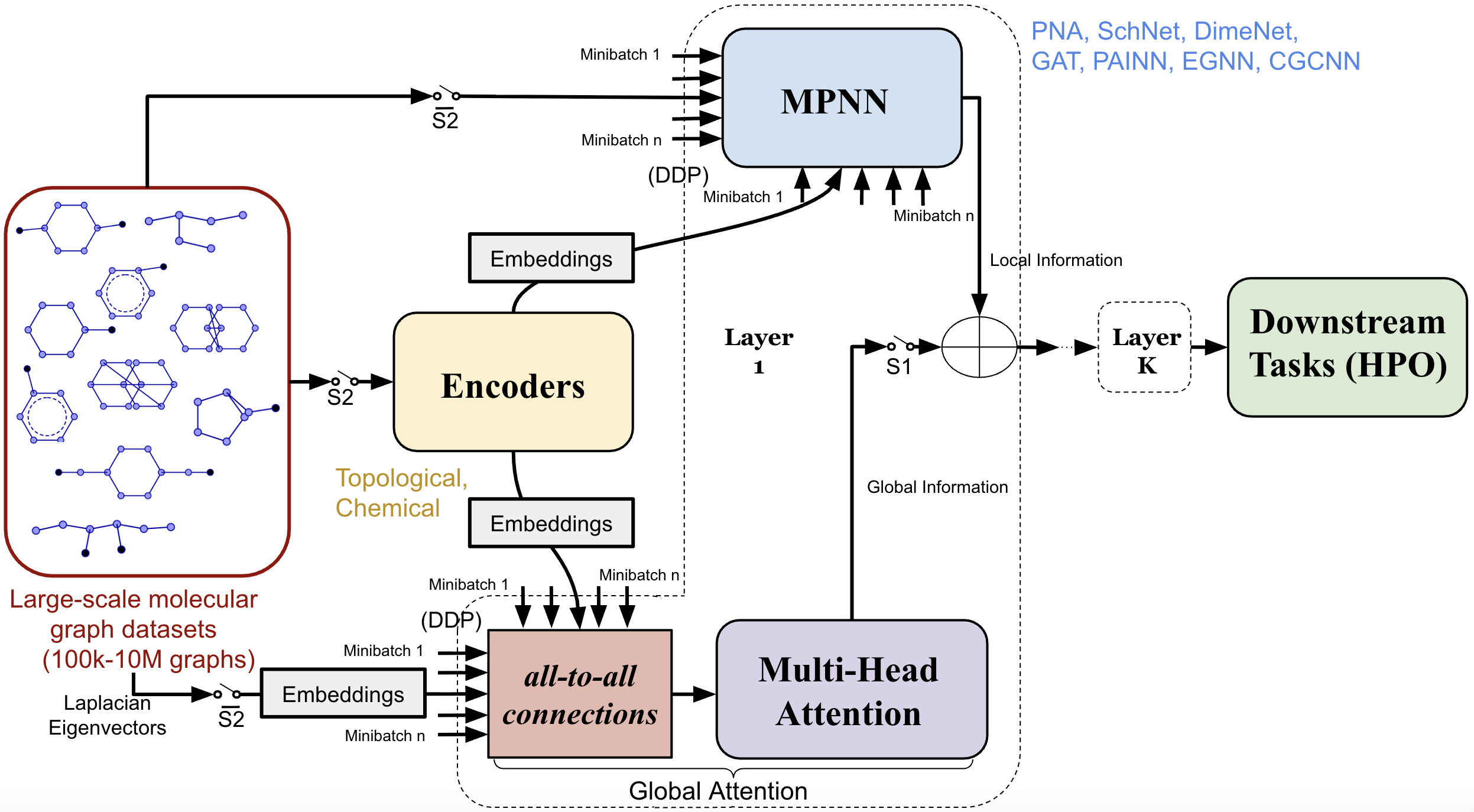}
    \caption{Flow diagram depicting the computational steps in our proposed framework. It depicts a $K-$layered model. \textbf{Each block operates parallelly on individual graphs}. Four independent pipelines are denoted by configurations of the switches S1 and S2. Ex., if S1 and S2 are both open,  HydraGNN pipeline is obtained. And, if S1 and S2 are both closed, global attention module is fused with local message-passing module while encoders provide domain-specific and positional information to both modules. Input features are embedded to suitable subspaces before feeding to the learnable modules. Output of layer $K$ is used for downstream tasks through hyperparameter optimization.}
    \label{fig:block}
\end{figure}

{We extended the existing HydraGNN framework (which implemented only MPNN mechanisms and that we still used as baseline for the results presented in this work) in order to integrate} \textit{MPNN and Global Attention}, thereby allows users {to adopt the upgraded HydraGNN framework} to seamlessly switch across independent model configurations that offer various techniques to aggregate information across graph data. Figure~\ref{fig:block} shows the different model configurations options provided by the newly expanded HydraGNN framework, toggled by switches S1 and S2.
\begin{enumerate}
    \item By opening both S1 and S2, the model resembles the HydraGNN~\citep{lupopasini2022hydragnn} architecture with its core functionalities such as DDP for large-scale datasets and a large selection of MPNN models -- both invariant and equivariant -- to choose from.~\label{scheme1}
    \item By closing S2 when S1 is open, we obtain an augmented version of HydraGNN where additional chemical and topological attributes are computed as a pre-processing step for each graph through an encoder module, which are first embedded using learnable transformations to a user-defined subspace and then leveraged by the MPNN module for the final downstream task.~\label{scheme2}
    \item Further, by opening S2 and closing S1, the model resembles the GPS~\citep{rampasek2022gps} architecture embedded within the HydraGNN pipeline. The suitably embedded \textit{laplacian eigenvectors} are used as positional encodings for the Global-Attention module. This scheme essentially combines the advantages of both models within a single end-to-end setup.~\label{scheme3}
    \item Finally, by closing both S1 and S2, we obtain a fully integrated model that combines the encoder, embedding, global attention and MPNN modules, together with DDP for large-scale data handling. We note that this model is in principle similar to GPS++~\citep{masters2023gpspp}. However, we do not use graphormer-style encodings and moreover, we use a broader class of MPNNs to aggregate local information than the aforementioned work.~\label{scheme4} 
\end{enumerate}
Crucial to the operation of this framework is the encoder module that pre-computes certain fundamental chemical properties of the molecules (at the level of the constituent atoms and the bonds) and topological properties of the corresponding graphs. Besides being useful for the Global-Attention module (that does not use the underlying local connectivity structure of the atomistic graph) to distinguish between nodes and edges of specific chemical types and topological structures and positions in the graph, these features also allow the MPNN modules to leverage the existing information that would otherwise be computed implicitly through message passing for a given downstream task. 

\subsection{Encoders}\label{sec:encoders}
Each atomistic graph is augmented with 3D coordinates of the constituent atoms, if the geometric information is available. The final model input for a graph consists of (i) raw atom/bond attributes, (ii) chemically informed per-atom descriptors, (iii) structural and positional encodings on nodes and edges, and (iv) simple geometric edge attributes. All feature blocks are standardized (zero mean, unit variance) per dataset split.

\noindent \textbf{Atoms.} For datasets that provide rich atom encodings (e.g., OGB~\citep{hu2020open} mol tasks), we adopt the standard $9$-dimensional atom feature vector covering atomic number, degree, formal charge, number of attached hydrogens, chirality, hybridization, aromaticity, and ring membership.\footnote{Edge features from OGB include bond type, stereochemistry, and conjugation.}
For QM9~\citep{ramakrishnan2014quantum,ruddigkeit2012enumeration} we use the atomic number $Z$ as the base node attribute, while for the NIAD MoF crystal dataset~\citep{burner2025abinitio} we use a two-channel node attribute $[Z,\ q^{(\mathrm{partial})}]$, where $q^{(\mathrm{partial})}$ is an (available) per-atom partial charge.

\noindent \textbf{Chemically informed descriptors (CE).} In addition to the raw attributes above, we compute a 15-dimensional per-atom descriptor vector using the following Mendeleev atomic properties: atomic weight, group, period, block (s/p/d/f), valence electron count, covalent radius, van der Waals radius, Pauling and Allen electronegativities, electron affinity, first ionization energy, melting point, boiling point, density, and atomic volume.  This descriptor matrix is standardized and concatenated as a separate channel available to the encoder.

\noindent \textbf{Topological encodings (TE).} We add two complementary, position- and structure-aware encodings:
(i) \textbf{Node PE}: a 9-feature stack of classic centrality/structure measures—degree, closeness, betweenness, eigenvector centrality, PageRank, local clustering coefficient, $k$-core number, harmonic centrality, and eccentricity—computed on the undirected atomistic graph and standardized.
(ii) \textbf{Edge PE}: a 4-feature vector per edge comprising edge betweenness, Jaccard coefficient, Adamic–Adar score, and preferential attachment, standardized.
In addition, we include \textbf{Laplacian eigenvector positional encodings (LPE)}: the top $k$ eigenvectors of the (symmetrized) graph Laplacian, which we compute once per graph (hyperparameter $k$ set in the config) and standardize.

\noindent \textbf{Bonds and geometry.} Bond features are customized based on the specific dataset used. For instance, OGB~\citep{hu2020open} supplies bond type, stereochemistry, conjugation information, while ZINC~\citep{irwin2005zinc} and QM9~\citep{ramakrishnan2014quantum,ruddigkeit2012enumeration} provide discrete bond attributes. 
For 3D structure, when coordinates are available we build the graph either from the dataset’s bond list or, for crystals, by a radius cutoff in Cartesian space, and attach the \emph{pairwise interatomic distance} $||\mathbf{r}_i-\mathbf{r}_j||$ as a continuous edge attribute. 
No explicit angle or dihedral features are used in our current experiments; instead, angular information is captured implicitly through message passing over the distance-augmented graph and the positional encodings.

Therefore, given a graph with $N$ nodes and $|\mathcal{E}|$ edges, the model consumes:
$\mathbf{X}\in\mathbb{R}^{N\times p}$ (raw atom features), 
$\mathbf{C}\in\mathbb{R}^{N\times 15}$ (Mendeleev descriptors), 
$\mathbf{P}\in\mathbb{R}^{N\times 9}$ (node topological encodings), 
$\mathbf{L}\in\mathbb{R}^{N\times d_l}$ (LPE), 
$\mathbf{E}\in\mathbb{R}^{|\mathcal{E}|\times f}$ (raw bond features and/or distances), and 
$\mathbf{G}\in\mathbb{R}^{M\times 4}$ (edge topological encodings).
Further, if the model is equivariant, it also consumes the 3D coordinates $\mathbf{R}\in\mathbb{R}^{M\times 3}$.
All channels are standardized before training, and samples with invalid values in computed encodings are discarded, which typically constitutes a negligible amount for each dataset.

\subsection{Embeddings}
\label{sec:feature-embedder}

We introduce a lightweight embedding module that unifies raw node/edge attributes with multiple classes of encodings and maps node and edge attributes to suitable subspaces of user-chosen dimensionality. Particularly, this module is employed for all schemes except the vanilla HydraGNN pipeline (Scheme~\ref{scheme1}), which operates directly on \textbf{X} and possibly \textbf{E}. It either embeds a combination of \textbf{X} and \textbf{L} on nodes and \textbf{E} on edges if only Global-Attention is activated (Scheme~\ref{scheme3}) while schemes~\ref{scheme2} and~\ref{scheme4} require embedding all the available encodings to a common subspace.

\noindent \textbf{Node embedding.} To build the input node embeddings for schemes~\ref{scheme2},~\ref{scheme3} and~\ref{scheme4} , we form a node input tensor by concatenation of the available encodings
\[
\mathbf{Z}_\text{node}
\;=\;
\big[\,\mathbf{X}\;\|\; \mathbf{L}\;\|\; \mathbf{P}\;\|\;\mathbf{C}\,\big]
\in\mathbb{R}^{N\times d_\text{node}^{\text{in}}},
\]
where absent tensors are simply omitted depending on the particular scheme being employed. A single linear projection with no bias maps to a hidden width $d_h$:
\[
\mathbf{H} \;=\; \mathbf{Z}_\text{node}\,\mathbf{W}_\text{node}, \qquad \mathbf{W}_\text{node}\in\mathbb{R}^{d_\text{node}^{\text{in}}\times d_h}.
\]
This preserves alignment across nodes while minimizing parameters. For Scheme~\ref{scheme1}, the module devolves to an identity on node encodings and returns $\mathbf{X}$ unchanged, enabling plug-and-play ablations.

\noindent \textbf{Edge embedding.} When using MPNNs that are desgined to handle edge features, we construct edge inputs via one of two modes: (1) $\mathbf{G}$ is used directly when encoders are employed or (2) if Global-Attention is activated but encoders are not used, for each edge $(u,v)$ with indices $(i,j)$, we compute the absolute difference of node-wise LPEs,
    \[
    \mathbf{r}_{(u,v)} \;=\; \big|\,\mathbf{L}_{i:} - \mathbf{L}_{j:}\,\big| \in \mathbb{R}^{d_\ell},
    \]
which is permutation-invariant and captures a simple notion of spectral displacement. Raw edge attributes $\mathbf{E}$, if available, are concatenated with the encodings. The resulting edge tensor
\[
\mathbf{Z}_\text{edge} \in \mathbb{R}^{E\times d_\text{edge}^{\text{in}}}
\]
is linearly projected (no bias) to an edge embedding $\mathbf{A}\in\mathbb{R}^{|\mathcal{E}|\times d_e'}$ and used to replace the input edge attributes $\mathbf{E} = \mathbf{A}$ to the model. For Scheme~\ref{scheme1}, the raw edge features are directly passed to the model. This lets downstream convolutions remain agnostic to the encoding sources while benefiting from unified edge representations.

In terms of complexity of the embedding operations, the dominant costs are two dense multiplications:
\[
\mathcal{O}\!\left(N\,d_\text{node}^{\text{in}}\,d_h\right) \quad\text{and}\quad
\mathcal{O}\!\left(|\mathcal{E}|\,d_\text{edge}^{\text{in}}\,d_e'\right),
\]
plus an $\mathcal{O}(|\mathcal{E}|\,d_\ell)$ LPE-difference when encoders are disabled. Since the MPNNs and Global-Attention already perform $\mathcal{O}(N^2)$ operations, the overall complexity of the model is unaffected by the additional embedding layers.
Further, we note the following main advantages of our method:
(i) \emph{Single-projection fusion.} Using one bias-free linear for nodes and edges each reduces parameters and avoids redundant per-channel MLPs while letting the backbone allocate depth where it matters. (ii) \emph{Decoupled toggles.} The four flags enable clean ablations: raw features only; raw+LPE; raw+TE(+CE); and edge-aware vs.\ node-only. (iii) \emph{Spectral relativity.} The LPE-difference path offers a cheap, sign-invariant relational signal when explicit RPEs are unavailable.

\subsection{Model Training}

Training is performed under the HydraGNN framework with hyperparameter optimization (HPO) driven by DeepHyper~\citep{balaprakash2018deephyper, egele2023asynchronous}. Let $\mathcal{H}$ denote the search space of hyperparameters (e.g., learning rate, hidden width $d_h$, edge width $d_e'$, dropout). For each candidate $h \in \mathcal{H}$, we train a model $f_{\theta,h}$ for $T$ epochs by minimizing the training loss
\[
\mathcal{L}(\theta;h) \;=\; \frac{1}{|\mathcal{D}_\text{train}|}\sum_{(x,y)\in \mathcal{D}_\text{train}} \ell\!\left(f_{\theta,h}(x), y\right),
\]
with $\ell$ typically chosen as squared error. Validation performance $\mathcal{L}_\text{val}(h)$ determines the best configuration
\[
h^\ast \;=\; \arg\min_{h\in\mathcal{H}} \; \mathcal{L}_\text{val}(h).
\]
We then re-train $f_{\theta,h^\ast}$ for an extended budget $T' \gg T$, subject to early stopping after $p$ consecutive epochs without improvement in $\mathcal{L}_\text{val}$. Model checkpointing ensures persistence of parameters $\theta_t$ at each epoch, providing both recovery and reproducibility.

The cost of the HPO phase scales as
\[
\mathcal{O}\!\big(|\mathcal{H}| \, T \, C_\text{epoch}\big),
\]
where $C_\text{epoch}$ denotes the cost of a single training epoch on the dataset. For graph data with $N$ nodes and $|\mathcal{E}|$ edges per sample, we approximate
\[
C_\text{epoch} = \mathcal{O}\!\left( \sum_{(x,y)\in\mathcal{D}_\text{train}} (N d_h^2 + |\mathcal{E}| d_e'^2) \right),
\]
where $d_h$ is the hidden node width and $d_e'$ the edge embedding width (cf.~Section~\ref{sec:feature-embedder}). After HPO, the re-training phase requires $\mathcal{O}(T' \, C_\text{epoch})$. 

\section{Experiments}\label{sec:exp}

\noindent Our experimental protocol evaluates the proposed framework across a broad suite of open-source atomistic datasets. For each dataset, we adopt the official train/validation/test split when provided; otherwise, we apply an \(80\%\!/\!10\%\!/\!10\%\) splitting. For every dataset and learning scheme, we perform hyperparameter optimization (HPO) to maximize the validation objective, select the best configuration, and then report that configuration on the held-out test set. Task-appropriate metrics are used throughout: mean absolute error (MAE) for regression targets, classification accuracy for OGB\mbox{-}PPA, and mean average precision (mAP) for OGB\mbox{-}PCBA. This procedure yields a standardized, reproducible comparison across datasets and provides an unbiased estimate of generalization to unseen graphs.

\subsection{Datasets}

\noindent We evaluate on \(7\) datasets spanning atomistic and biochemical graph regimes (Table~\ref{tab:datasets}). {We intentionally use widely available, community-standard datasets to ensure reproducibility and to enable controlled cross-architecture comparisons under identical HPO/training settings, rather than focusing on application-specific proprietary datasets.} The collection comprises \(\sim 4.21\)M graphs with mean sizes from \(\sim 18\) to \(\sim 243\) nodes, covering small molecules (QM9, ZINC, OGB\mbox{-}PCQM) and larger, more complex structures (TMQM, NIAID, OGB\mbox{-}PPA). Supervision includes four graph\mbox{-}level regression targets—free energy at \(298.15\,\mathrm{K}\) (QM9), constrained solubility (ZINC), dispersion energy (TMQM), and the HOMO--LUMO gap (OGB\mbox{-}PCQM)—one node\mbox{-}level regression target (partial charges in NIAID), and two graph\mbox{-}level classification benchmarks: multi\mbox{-}class accuracy on OGB\mbox{-}PPA and multi\mbox{-}label mAP on OGB\mbox{-}PCBA. 

We set the LPE dimensionality \(d_\ell\) to the number of smallest nonzero Laplacian eigenvectors that are consistently available across graphs in each dataset; a small fraction of graphs with fewer than \(d_\ell\) nontrivial eigenvectors are omitted. Table~\ref{tab:datasets} lists \(d_\ell\) per dataset.


\begin{table}[h!]
\centering
\caption{Graph dataset summary and LPE dimensionality.}
\begin{tabular}{@{}llll@{}}
\toprule
\textbf{Dataset} & \textbf{\shortstack{Avg.\\\# Nodes}} & \textbf{\shortstack{\#\\Graphs}} & \textbf{\shortstack{\#\\lapPE (\(d_\ell\))}} \\
\midrule
QM9         & $\sim$18     & 130k    &  2 \\
ZINC        & $\sim$23     & 250k    &  5 \\
TMQM        & $\sim$120    & 100k    &  6 \\
NIAID       & $\sim$200    & 130k    &  6 \\
OGB-PCQM4Mv2    & $\sim$20     & 3M      &  4 \\
OGB-PPA     & $\sim$243    & 158k    &  5 \\
OGB-molPCBA    & $\sim$26     & 438k    &  5 \\
\bottomrule
\end{tabular}
\label{tab:datasets}
\end{table}

\noindent \textbf{QM9} provides equilibrium geometries and a standardized suite of quantum\mbox{-}chemical properties for small organic molecules curated from the GDB chemical universe \cite{qm9_sdata2014,gdb17_jcim2012}. Nodes encode atomic number; edges correspond to covalent bonds with one\mbox{-}hot bond type (single/double/triple/aromatic). QM9 offers \(19\) regression targets: {Dipole moment}, {Isotropic polarizability}, {Highest occupied molecular orbital energy}, {Lowest unoccupied molecular orbital energy}, {Gap between HOMO and LUMO}, {Electronic spatial extent}, {Zero point vibrational energy}, {Internal energy at 0K}, {Internal energy at 298.15K}, {Enthalpy at 298.15K}, {Free energy at 298.15K}, {Heat capacity at 298.15K}, {Atomization energy at 0K}, {Atomization energy at 298.15K}, {Atomization enthalpy at 298.15K}, {Atomization free energy at 298.15K}, {Rotational constant A}, {Rotational constant B}, {Rotational constant C}. In our experiments, we predict \texttt{Free energy at 298.15K}. \\

\noindent \textbf{ZINC} is a freely available repository of purchasable small molecules prepared in 3D, protonation, and tautomeric states for virtual screening; we reference both the original release and ZINC15 \cite{zinc2005,zinc15_2015}. Nodes represent heavy atoms with features including atomic number (atom type), chirality, atomic degree, formal charge, number of attached hydrogens, number of radical electrons, hybridization state, aromaticity, and ring membership. Edges are covalent bonds with bond order (single/double/triple/aromatic), stereo configuration (e.g., cis/trans), and a binary conjugation indicator. The task is to regress over graph\mbox{-}level \texttt{Constrained Solubility} (logP). \\

\noindent \textbf{TMQM} comprises quantum\mbox{-}optimized geometries and electronic properties for a large, structurally diverse set of mononuclear transition\mbox{-}metal complexes from the Cambridge Structural Database \cite{tmqm_jcim2020}. We construct geometric graphs using a \(5.0\)\AA\ radius cutoff. Node features include atomic number, formal charge, atomic valence indices, and 3D coordinates; edges carry real\mbox{-}valued Euclidean distances for atom pairs within the cutoff. The dataset provides \(11\) targets (e.g., molecular charge, spin, metal coordination degree, electronic energy, dispersion energy, dipole moment, natural charge at the metal center, HOMO--LUMO gap, HOMO/LUMO energies, polarizability); here we focus on graph\mbox{-}level \texttt{Dispersion Energy} regression. \\

\noindent \textbf{NIAID} contains experimental and computational metal\mbox{-}organic frameworks with REPEAT DFT\mbox{-}derived partial charges and precomputed descriptors, accessed via the NIAID Data Discovery Portal \cite{arc_mof_zenodo2022,arc_mof_chemrxiv2022}. Graphs are built with a \(5.0\)\AA\ radius cutoff. Node features are atomic numbers; edges encode Euclidean distances. The task is node\mbox{-}level regression of per\mbox{-}atom \texttt{Partial Charges}. \\

\noindent \textbf{OGB\mbox{-}PCQM (PCQM4Mv2)} defines prediction of the DFT \texttt{HOMO-LUMO gap} from 2D molecular graphs curated from PubChemQC, with standardized splits and evaluation protocols for large\mbox{-}scale graph learning \cite{pcqm4mv2_doc,ogb_lsc_2021,pubchemqc_jcim2017}. \\

\noindent \textbf{OGB\mbox{-}PCBA} is OGB’s adaptation of the MoleculeNet molPCBA suite derived from PubChem BioAssay, providing many binary endpoints and strong baselines for evaluating classifier calibration and transfer \cite{ogb_2020,moleculenet2018}. We use this dataset for graph\mbox{-}level multi\mbox{-}label classification. \\

\noindent \emph{Featurization for OGB\mbox{-}PCQM and OGB\mbox{-}PCBA.} Both datasets share the same molecular featurization. Each node is annotated with its atomic number, chirality type (e.g., R/S stereocenter), atomic degree, formal charge, number of attached hydrogens, number of radical electrons, hybridization state, an aromaticity flag, and a ring\mbox{-}membership indicator. Edges correspond to covalent bonds and carry bond type (single/double/triple/aromatic), bond stereo configuration (e.g., cis/trans), and a binary indicator for conjugation. \\

\noindent \textbf{OGB\mbox{-}PPA} consists of protein–protein association (PPA) network neighborhoods spanning many species and taxonomic groups; graphs are derived from STRING associations and labels reflect the source group, enabling evaluation of biological graph generalization \cite{ogb_graphprop_doc,string_v11_2019}. We use this dataset for protein\mbox{-}graph classification (37\mbox{-}class, evaluated by accuracy). Nodes represent proteins and do not carry intrinsic atom\mbox{-}level features. Edges encode associations; in our setup, each edge includes a 7\mbox{-}dimensional binary vector indicating evidence channels (e.g., gene co\mbox{-}occurrence, gene fusion events, co\mbox{-}expression), with \(1\) denoting the presence of the corresponding evidence type.

\subsection{Hyperparameter search space (HPO)}

We specify a conditional search space over MPNNs and global attention. The space branches on (i) availability of 3D coordinates (\texttt{has\_pos}), (ii) whether a global attention block is enabled (\texttt{global\_attn\_engine}), and (iii) whether feature encodings are used (\texttt{use\_encodings}). Discrete ranges denote integer-valued grids; braces \(\{\cdot\}\) enumerate categorical options.

\begin{table}[h!]
\centering
\caption{Model family by coordinate availability.}
\begin{tabular}{@{}ll@{}}
\toprule
\textbf{Condition} & \textbf{\texttt{mpnn\_type}} options \\ \midrule
\texttt{has\_pos} = True  & \{\texttt{PNA}, \texttt{CGCNN}, \texttt{SchNet}, \texttt{DimeNet}, \texttt{EGNN},   \texttt{PAINN}\} \\
\texttt{has\_pos} = False & \{\texttt{GAT}, \texttt{GINE}, \texttt{PNA}, \texttt{CGCNN}\} \\
\bottomrule
\end{tabular}
\label{tab:hpo-model-families}
\end{table}

\noindent Non\mbox{-}equivariant architectures (e.g., \texttt{PNA}, \texttt{CGCNN}) do not consume 3D coordinates (\texttt{pos}) and therefore operate identically on 2D graphs. We nevertheless include them in the HPO search space alongside equivariant models (as shown in Table~\ref{tab:hpo-model-families}, \texttt{has\_pos} = True ) to enable data\mbox{-}driven selection and to quantify the marginal benefit of explicit geometric information -- and SE(3)\mbox{-}equivariance -- over standard message passing.

\begin{table}[h!]
\centering
\caption{Search space when \texttt{global\_attn\_engine} = False.}
\begin{tabular}{@{}lll@{}}
\toprule
\textbf{Hyperparameter} & \textbf{\texttt{use\_encodings}=False} & \textbf{\texttt{use\_encodings}=True} \\ \midrule
\texttt{num\_conv\_layers}  & \(\{1,2,3,4,5,6\}\) & \(\{1,2,3,4,5,6\}\) \\
\texttt{global\_attn\_heads} & \(\{0\}\) & \(\{0\}\) \\
\texttt{hidden\_dim}        & [4,32] & [16,64] \\
\texttt{edge\_embed\_dim}   & \(\{0\}\) & \(\{0,4,5,6,7,8,9,10,11,12\}\) \\
\bottomrule
\end{tabular}
\label{tab:hpo-attn-off}
\end{table}

\begin{table}[h!]
\centering
\caption{Search space when \texttt{global\_attn\_engine} = True.}
\begin{tabular}{@{}lll@{}}
\toprule
\textbf{Hyperparameter} & \textbf{\texttt{use\_encodings}=False} & \textbf{\texttt{use\_encodings}=True} \\ \midrule
\texttt{num\_conv\_layers}  & \(\{1,2,3\}\) & \(\{1,2,3\}\) \\
\texttt{global\_attn\_heads} & \(\{2,4,8\}\) & \(\{2,4,8\}\) \\
\texttt{hidden\_dim}        & \(\{8,16,24,32,40,48\}\) & \(\{16,24,32,40,48,56,64\}\) \\
\texttt{edge\_embed\_dim}   & \(\{0,4,5,6,8,9,10,11,12\}\) & \(\{0,4,5,6,7,8,9,10,11,12\}\) \\
\bottomrule
\end{tabular}
\label{tab:hpo-attn-on}
\end{table}

\noindent \noindent Multi–head attention imposes the constraint \(\texttt{hidden\_dim} \,\%\, \texttt{global\_attn\_heads}=0\), i.e., the per–head width \(d_{\text{head}}=\texttt{hidden\_dim}/\texttt{global\_attn\_heads}\) must be an integer. Accordingly, in Table~\ref{tab:hpo-attn-on} we restrict \texttt{hidden\_dim} to multiples of \(\mathrm{lcm}(2,4,8)=8\), ensuring every listed \texttt{hidden\_dim} is compatible with all admissible values of \texttt{global\_attn\_heads} and preventing invalid HPO configurations. \noindent When \texttt{global\_attn\_engine} is enabled, we widen the candidate set for \texttt{hidden\_dim} to admit larger layer widths. This reflects the increased effective receptive field and information aggregation introduced by global attention, allowing the model to be more expressive. \\

\noindent\emph{Equivariance toggle.} Setting \(\texttt{edge\_embed\_dim}=0\) uses raw interatomic distance as the sole edge attribute, preserving SE(3)-equivariance for equivariant architectures (e.g., \texttt{SchNet}, \texttt{DimeNet}, \texttt{EGNN},  \texttt{PAINN}). Choosing \(\texttt{edge\_embed\_dim}>0\) activates the full edge-feature pipeline (e.g., bond/type/auxiliary encodings when available), which generally breaks strict equivariance but can increase expressivity. For the non-equivariant models (\texttt{GAT, GINE, PNA} and \texttt{CGCNN}), \(\texttt{edge\_embeb\_dim}=0\) allows the model to use the raw edge features directly without learnable projections.

\subsection{Hardware Setting}
All numerical experiments were conducted on the OLCF Frontier supercomputer, the first U.S. exascale system. Frontier is based on the HPE Cray EX architecture and consists of 9,472 compute nodes, each equipped with one AMD EPYC™ 7A53 64-core CPU and four AMD Instinct™ MI250X GPUs. Each node offers a high-bandwidth memory subsystem with 512 GB of DDR4 memory on the CPU and 128 GB of HBM2e memory across the GPUs, connected via AMD’s Infinity Fabric. Nodes are linked through the HPE Slingshot-11 interconnect, which provides high bandwidth and low latency for large-scale parallel applications. The system is supported by a multi-petabyte Lustre-based parallel file system to enable high-throughput I/O for large-scale simulations.

\subsection{Numerical Results}

We evaluate four training schemes on all our datasets: (S1) \texttt{GPS} disabled, encoders disabled{, which corresponds exactly to the original HydraGNN architecture}; (S2) \texttt{GPS} disabled, encoders enabled; (S3) \texttt{GPS} enabled, encoders disabled; (S4) \texttt{GPS} enabled, encoders enabled. For each scheme we report the best HPO trial and its test performance (MSE/MAE/Pearson $r$ as indicated in the plot titles).


\subsubsection{ZINC}\label{sec:zinc}

\begin{figure}[!h]
  \centering
  \begin{subfigure}{0.48\textwidth}
    \centering
    \includegraphics[width=\linewidth]{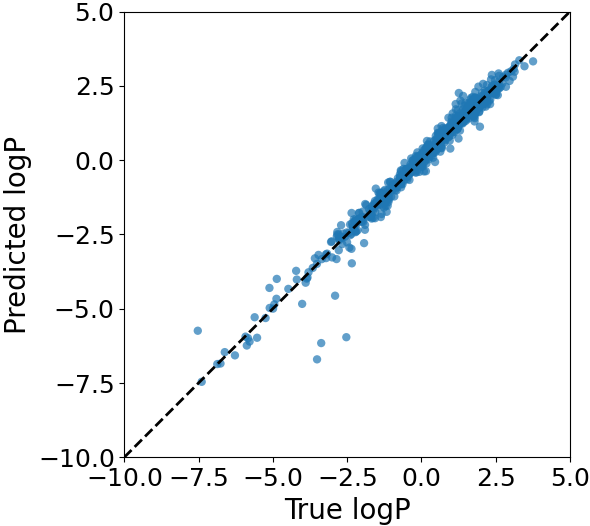}
    \caption{S1: GPS and Encoders disabled}
    \label{fig:zinc_a}
  \end{subfigure}\hfill
  \begin{subfigure}{0.48\textwidth}
    \centering
    \includegraphics[width=\linewidth]{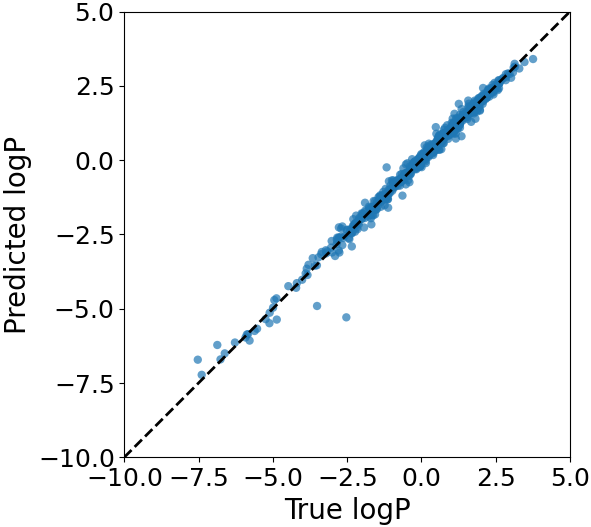}
    \caption{S2: GPS disabled, Encoders enabled}
    \label{fig:zinc_b}
  \end{subfigure}

  \medskip

  \begin{subfigure}{0.48\textwidth}
    \centering
    \includegraphics[width=\linewidth]{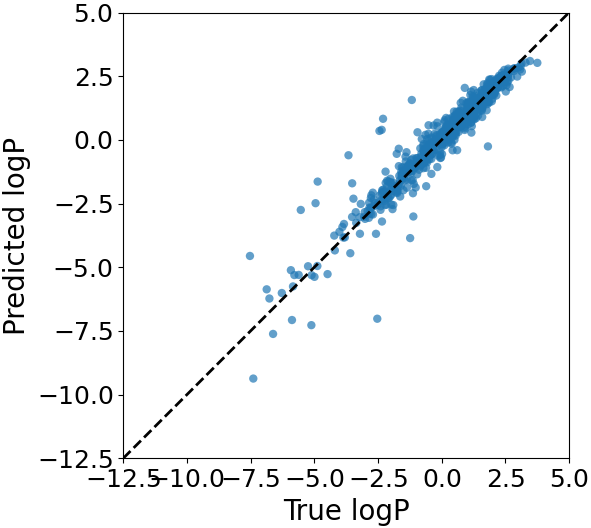}
    \caption{S3: GPS enabled, Encoders disabled}
    \label{fig:zinc_c}
  \end{subfigure}\hfill
  \begin{subfigure}{0.48\textwidth}
    \centering
    \includegraphics[width=\linewidth]{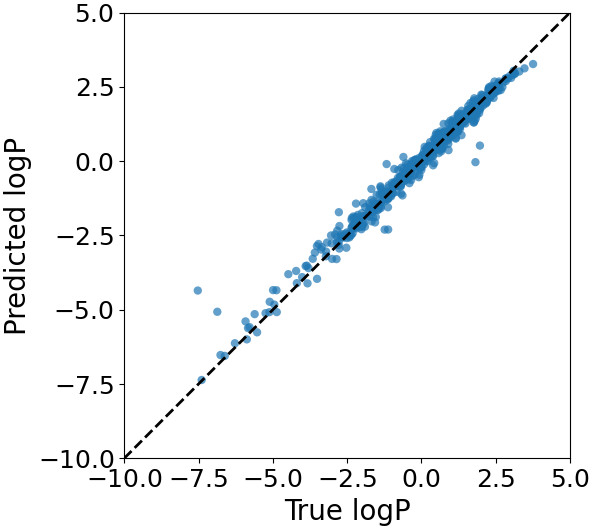}
    \caption{S4: GPS and Encoders enabled}
    \label{fig:zinc_d}
  \end{subfigure}
  \caption{ZINC parity plots (Predicted vs.\ True logP) for the best HPO trial per scheme }
  \label{fig:zinc-parity-4up}
\end{figure}

Tables~\ref{tab:zinc-hpo} and~\ref{tab:zinc-metrics} indicate that enabling learned encoders without GPS (\textbf{S2}) yields the most favorable bias–variance trade-off for this dataset. 
Relative to the non-encoder baseline (\textbf{S1}), MSE decreases by \(\sim62\%\) and MAE by \(\sim37\%\) at a moderate parameter increase ($185$k vs.\ $111$k). 
In contrast, activating GPS without encoders (\textbf{S3}) reduces parameter count to $62$k but substantially degrades accuracy, suggesting that additional global information cannot compensate for weaker token features. 
Adding GPS on top of encoders (\textbf{S4}, $207$k parameters) recovers much of the performance lost by \textbf{S3} yet remains inferior to \textbf{S2}, implying diminishing returns from global information. 
The parity plots (Figure~\ref{fig:zinc-parity-4up}) show tightly clustered residuals with a small number of outliers; \textbf{S2} exhibits the highest similarity to the identity with the least dispersion in the high-magnitude regime.

\begin{table}[!h]
\centering
\caption{ZINC best HPO configuration per scheme}
\label{tab:zinc-hpo}
\begin{tabular}{@{}lcccccccc@{}}
\toprule
\textbf{Scheme} & \textbf{MPNN} & \textbf{\#Conv} & \textbf{Hidden} & \textbf{EdgeEmb} & \textbf{GPS Heads} & \textbf{\#Parameters} \\
\midrule
S1  & PNA & 6 & 32 & 0  & 0 & 111388\\
S2 & PNA & 4 & 46 & 4  & 0 & 185318\\
S3 & PNA & 2 & 32 & 9  & 4 & 62374\\
S4 & PNA & 3 & 48 & 10 & 8 & 207288\\
\bottomrule
\end{tabular}
\end{table}

\begin{table}[!h]
\centering
\caption{ZINC test performance by scheme}
\label{tab:zinc-metrics}
\begin{tabular}{@{}lcccc@{}}
\toprule
\textbf{Scheme} & \textbf{MSE} $\downarrow$ & \textbf{MAE} $\downarrow$ & \textbf{Pearson $r$} $\uparrow$ \\
\midrule
S1     & 0.112859 & 0.197555 & 0.985365 \\
S2    & \textbf{0.042459} & \textbf{0.125066} & \textbf{0.994298}  \\
S3   & 0.316814 & 0.326824 & 0.957194 \\
S4  & 0.083230 & 0.181921 & 0.989287 \\
\bottomrule
\end{tabular}
\end{table}

\subsubsection{QM9}\label{sec:qm9}

\begin{figure*}[!h]
  \centering
  \begin{subfigure}[t]{0.48\textwidth}
    \centering
    \includegraphics[width=\linewidth]{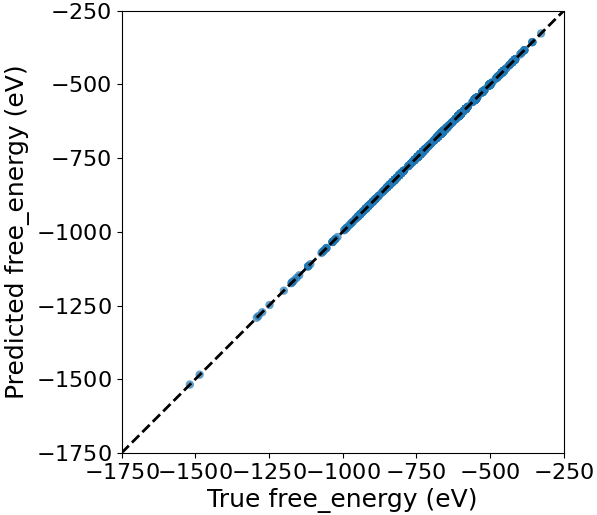}
    \caption{S1: GPS and Encoders disabled}
  \end{subfigure}\hfill
  \begin{subfigure}[t]{0.48\textwidth}
    \centering
    \includegraphics[width=\linewidth]{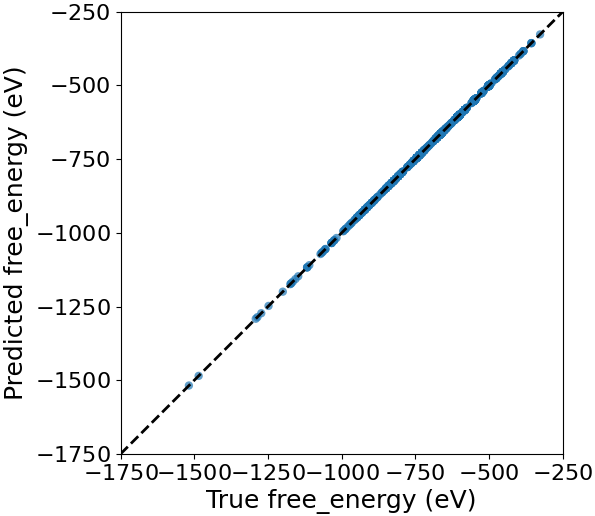}
    \caption{S2: GPS disabled, Encoders enabled}
  \end{subfigure}

  \vspace{0.6em}
  \begin{subfigure}[t]{0.48\textwidth}
    \centering
    \includegraphics[width=\linewidth]{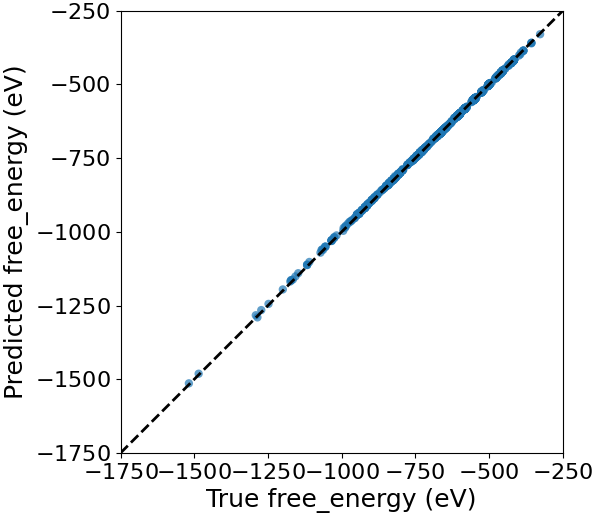}
    \caption{S3: GPS enabled, Encoders disabled}
  \end{subfigure}\hfill
  \begin{subfigure}[t]{0.48\textwidth}
    \centering
    \includegraphics[width=\linewidth]{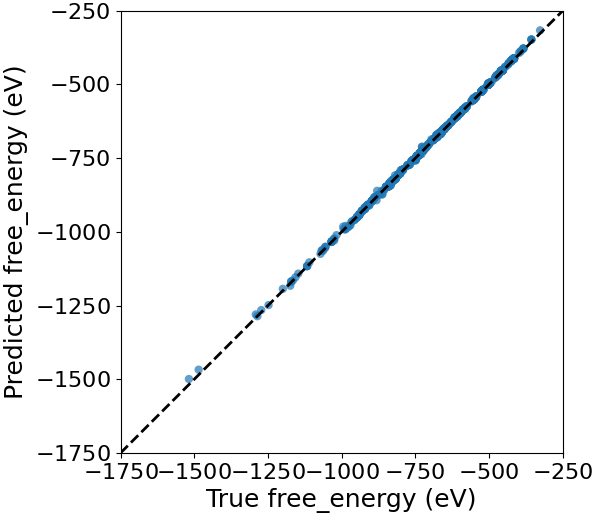}
    \caption{S4: GPS and Encoders enabled}
  \end{subfigure}
  \caption{QM9 parity plots (Predicted vs.\ True free energy) for the best HPO trial per scheme}
  \label{fig:qm9-freeenergy-scatter}
\end{figure*}

In this case, all schemes achieve \(r\approx 1\) (Figure~\ref{fig:qm9-freeenergy-scatter}), yet error magnitudes depend strongly on architectural choices (Tables~\ref{tab:qm9-schemes} and~\ref{tab:qm9-metrics}). The encoder-augmented PAINN without GPS (\textbf{S2}, $82.9$k parameters) attains the lowest MSE/MAE, improving upon the compact PAINN baseline (\textbf{S1}, $15.1$k) by \(\sim20\%\) MSE and \(\sim9\%\) MAE; thus \textbf{S1} is highly parameter-efficient but \textbf{S2} achieves the best absolute error. GPS-enabled variants (\textbf{S3} DimeNet, \textbf{S4} PAINN) exhibit markedly larger errors despite near-perfect correlations, consistent with a calibration bias rather than rank ordering failures—visible as a slight slope offset from the identity in Figure~\ref{fig:qm9-freeenergy-scatter}. The results suggest that for QM9 the quality of local information and the hidden layer dimension plays an important role, whereas shallow global attention (single-layered) confers no benefit at comparable or even lower model sizes.

\begin{table}[!h]
  \centering
  \caption{QM9 best HPO configuration per scheme}
  \label{tab:qm9-schemes}
  \begin{tabular}{l
                  S[table-format=2.0]
                  S[table-format=2.0]
                  S[table-format=1.0]
                  S[table-format=2.0]
                  S[table-format=1.0]
                  S[table-format=1.0]
                  S[table-format=1.0]}
    \toprule
    \textbf{Scheme} & \textbf{MPNN} & \textbf{\#Conv} & \textbf{Hidden} & \textbf{EdgeEmb} & \textbf{GPS Heads} & \textbf{\#Parameters} \\
    \midrule
    S1 & PAINN  & 2 & 15 & 0  & 0 & 15119\\
    S2  & PAINN  & 2 & 47 & 12  & 0 & 82898\\
    S3  & \text{DimeNet} & 2 &  8 & 4 & 2 & 10948\\
    S4 & PAINN & 1 & 20 & 10  & 2 & 17992 \\
    \bottomrule
  \end{tabular}
\end{table}

\begin{table}[!h]
  \centering
  \caption{QM9 test performance by scheme}
  \label{tab:qm9-metrics}
  \sisetup{round-mode=places,round-precision=6}
  \begin{tabular}{l
                  S[table-format=2.6]
                  S[table-format=1.6]
                  S[table-format=1.6]}
    \toprule
    \textbf{Scheme} & \textbf{MSE} $\downarrow$ & \textbf{MAE} $\downarrow$ & \textbf{Pearson $r$} $\uparrow$ \\
    \midrule
    S1 & 0.002416 & 0.036195 & {\bfseries 1.000000} \\
    S2 & {\bfseries 0.001934} & {\bfseries 0.032784} & {\bfseries 1.000000} \\
    S3 & 4.415277 & 1.551903 & 0.999958 \\
    S4 & 7.484828 & 2.115828 & 0.999893 \\
    \bottomrule
  \end{tabular}
\end{table}

\subsubsection{TMQM}\label{sec:tmqm}

\begin{figure*}[!h]
  \centering
  \begin{subfigure}[t]{0.48\textwidth}
    \centering
    \includegraphics[width=\linewidth]{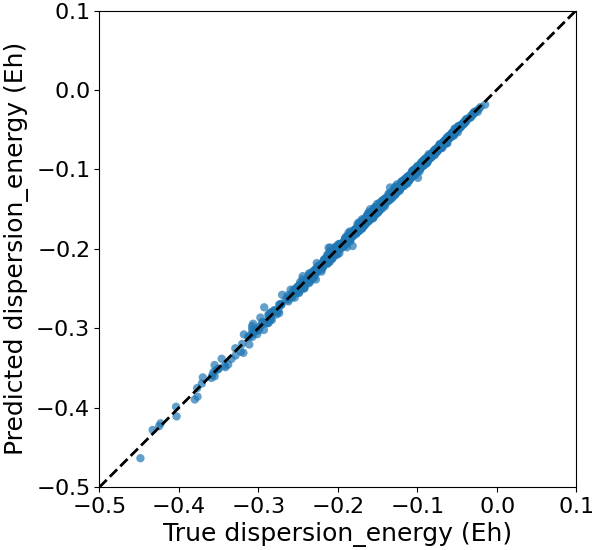}
    \caption{S1: GPS and Encoders disabled}
  \end{subfigure}\hfill
  \begin{subfigure}[t]{0.48\textwidth}
    \centering
    \includegraphics[width=\linewidth]{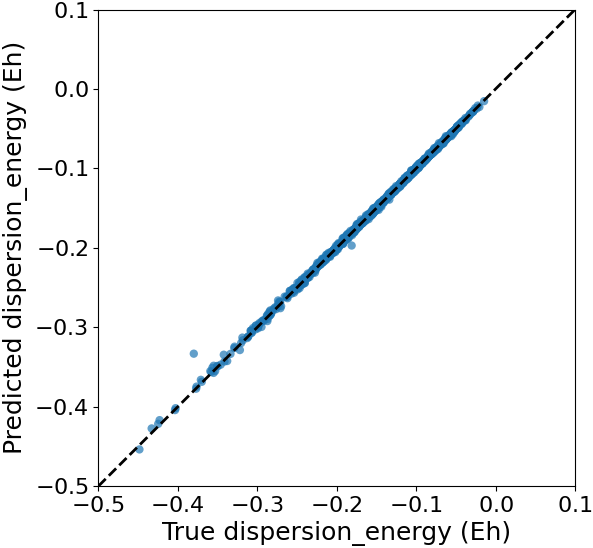}
    \caption{S2: GPS disabled, Encoders enabled}
  \end{subfigure}

  \vspace{0.6em}
  \begin{subfigure}[t]{0.48\textwidth}
    \centering
    \includegraphics[width=\linewidth]{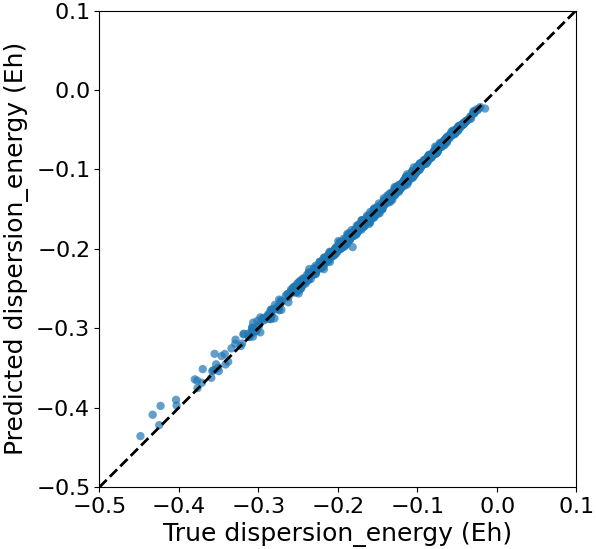}
    \caption{S3: GPS enabled, Encoders disabled}
  \end{subfigure}\hfill
  \begin{subfigure}[t]{0.48\textwidth}
    \centering
    \includegraphics[width=\linewidth]{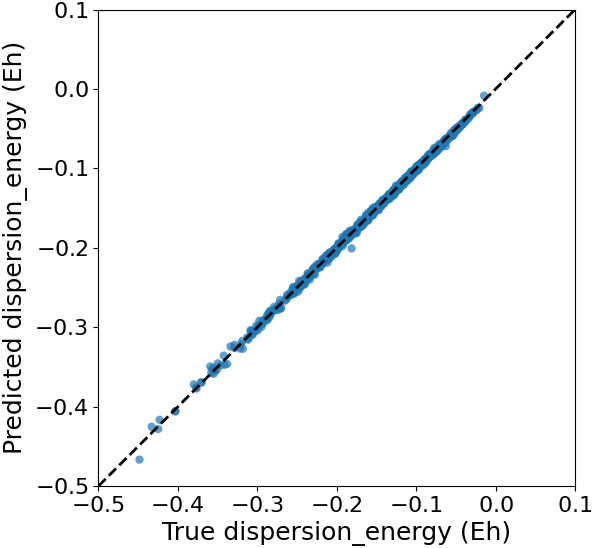}
    \caption{S4: GPS and Encoders enabled}
  \end{subfigure}
  \caption{TMQM parity plots (Predicted vs.\ True dispersion energy) for the best HPO trial per scheme. }
  \label{fig:tmqm-dispersion-scatter}
\end{figure*}

For this dataset, all configurations lead to similar high quality results (Figure~\ref{fig:tmqm-dispersion-scatter}), and variations in performance across configurations are very small (Table~\ref{tab:tmqm-metrics}). Notably, the best model (\textbf{S2}, PAINN, $63.7$k parameters, $1$ conv layer) outperforms a substantially larger GPS variant (\textbf{S3}, $184.6$k) and even a deeper PAINN (\textbf{S1}, $65.9$k), indicating that increased depth and global heads are unnecessary once sufficiently expressive local encoders are present. The smallest model (\textbf{S4}, $27.5$k) is close to \textbf{S2}, underscoring that this target is amenable to compact networks; additional capacity mainly yields marginal gains.

\begin{table}[!h]
  \centering
  \caption{TMQM best HPO configuration per scheme}
  \label{tab:tmqm-schemes}
  \sisetup{table-number-alignment = center}
  \begin{tabular}{l
                  S[table-format=2.0]
                  S[table-format=2.0]
                  S[table-format=1.0]
                  S[table-format=2.0]
                  S[table-format=1.0]
                  S[table-format=1.0]
                  S[table-format=1.0]}
    \toprule
\textbf{Scheme} & \textbf{MPNN} & \textbf{\#Conv} & \textbf{Hidden} & \textbf{EdgeEmb} & \textbf{GPS Heads} & \textbf{\#Parameters}\\
    \midrule
    S1 &  PNA  & 4 & 30 & 0   & 0 & 65933\\
    S2 &  PAINN & 1 & 59 & 12 & 0 & 63738\\
    S3 &  PAINN & 3 & 48 & 11 & 8 & 184639\\
    S4 &  PAINN & 3 & 16 & 12 & 8 & 27486\\
    \bottomrule
  \end{tabular}
\end{table}

\begin{table}[!h]
  \centering
  \caption{TMQM test performance per scheme}
  \label{tab:tmqm-metrics}
  \sisetup{round-mode=places,round-precision=6}
  \begin{tabular}{l
                  S[table-format=1.6]
                  S[table-format=1.6]
                  S[table-format=1.6]
                  S[table-format=1.6]}
    \toprule
    \textbf{Scheme} & \textbf{MSE} $\downarrow$ & \textbf{MAE} $\downarrow$ & \textbf{Pearson $r$} $\uparrow$ \\
    \midrule
    S1 & 0.000008 & 0.002009 & 0.999169 \\
    S2 & \textbf{0.000004} & \textbf{0.001135} & \textbf{0.999598} \\
    S3 & 0.000010 & 0.002055 & 0.999090 \\
    S4 & 0.000006 & 0.001843 & 0.999560 \\
    \bottomrule
  \end{tabular}
\end{table}

\subsubsection{NIAID}\label{sec:niaid}

\begin{figure*}[!h]
  \centering
  \begin{subfigure}[t]{0.48\textwidth}
    \centering
\includegraphics[width=\linewidth,height=\linewidth,keepaspectratio]{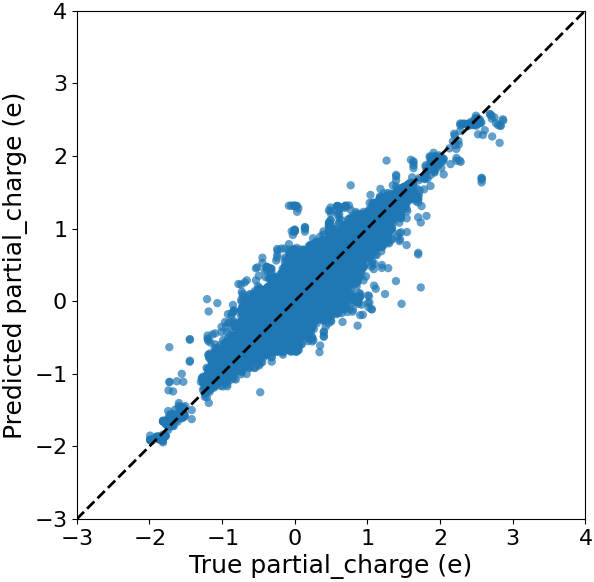}
    \caption{S1: GPS and Encoders disabled}
  \end{subfigure}\hfill
  \begin{subfigure}[t]{0.48\textwidth}
    \centering
    \includegraphics[width=\linewidth]{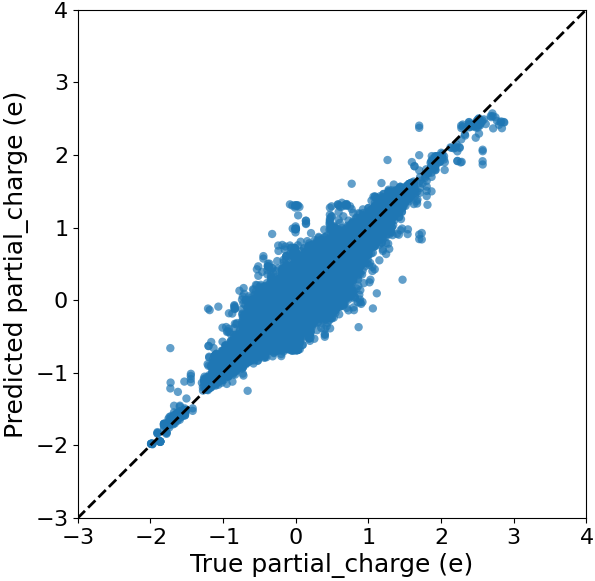}
    \caption{S2: GPS disabled, Encoders enabled}
  \end{subfigure}

  \vspace{0.6em}
  \begin{subfigure}[t]{0.48\textwidth}
    \centering
    \includegraphics[width=\linewidth]{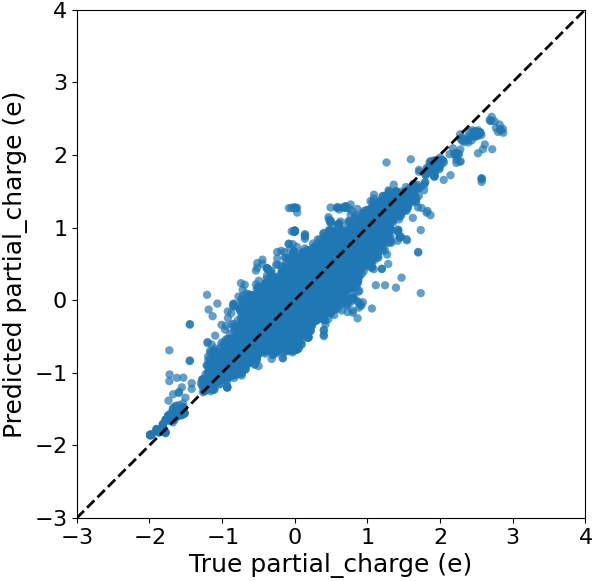}
    \caption{S3: GPS enabled, Encoders disabled}
  \end{subfigure}\hfill
  \begin{subfigure}[t]{0.48\textwidth}
    \centering
    \includegraphics[width=\linewidth]{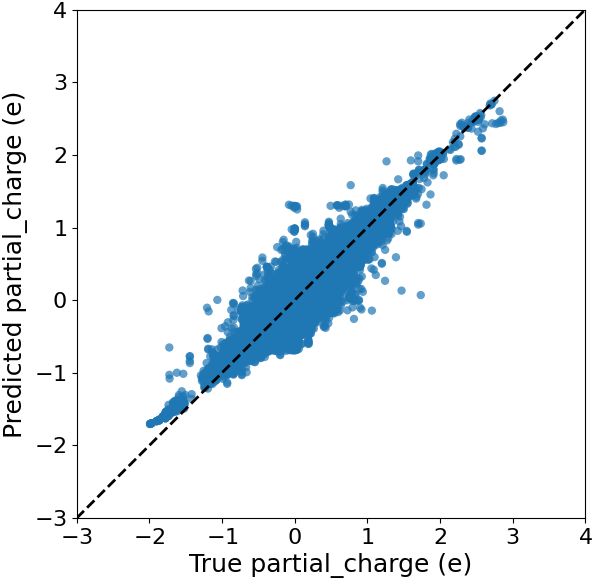}
    \caption{S4: GPS and Encoders enabled}
  \end{subfigure}
  \caption{NIAID parity plots (Predicted vs.\ True partial charge) for the best HPO trial per scheme}
  \label{fig:niaid-mof-partialcharge-scatter}
\end{figure*}

In this case, GPS provides modest but consistent improvements when paired with adequate width (Tables~\ref{tab:niaid-mof-schemes} and~\ref{tab:niaid-mof-metrics}). \textbf{S4} achieves the best MSE and highest correlation, while \textbf{S2} ($282.6$k) attains the lowest MAE; both clearly outperform the smaller no-GPS baseline (\textbf{S1}, $98.3$k). The gap between \textbf{S4} and \textbf{S2} is small (e.g., \(\sim1.8\%\) MSE), implying diminishing returns per parameter, yet the parity plots (Figure~\ref{fig:niaid-mof-partialcharge-scatter}) show a slightly tighter envelope for \textbf{S4} across the full dynamic range, consistent with GPS aiding long-range charge redistribution effects. Deeper stacks are not required ($3–4$ conv layers suffice), but larger hidden/edge embeddings appear beneficial.

\begin{table}[!h]
  \centering
  \caption{NIAID best HPO configuration per scheme}
  \label{tab:niaid-mof-schemes}
  \sisetup{table-number-alignment = center}
  \begin{tabular}{l
                  S[table-format=2.0]
                  S[table-format=2.0]
                  S[table-format=1.0]
                  S[table-format=2.0]
                  S[table-format=1.0]
                  S[table-format=1.0]
                  S[table-format=1.0]}
    \toprule
    \textbf{Scheme} & \textbf{MPNN} & \textbf{\#Conv} & \textbf{Hidden} & \textbf{EdgeEmb} & \textbf{GPS Heads} & \textbf{\#Parameters} \\
    \midrule
    S1 &  PAINN & 6 & 32 & 0 & 0 &  98256 \\
    S2 &  PAINN & 4 & 63 & 8 & 0 & 282600\\
    S3 &  PAINN & 3 & 48 & 5 & 4 & 178573\\
    S4 &  PAINN & 3 & 64 & 12 & 4 & 316774\\
    \bottomrule
  \end{tabular}
\end{table}

\begin{table}[!h]
  \centering
  \caption{NIAID test performance per scheme}
  \label{tab:niaid-mof-metrics}
  \sisetup{round-mode=places,round-precision=6}
  \begin{tabular}{l
                  S[table-format=1.6]
                  S[table-format=1.6]
                  S[table-format=1.6]}
    \toprule
    \textbf{Scheme} & \textbf{MSE} $\downarrow$ & \textbf{MAE} $\downarrow$ & \textbf{Pearson $r$} $\uparrow$ \\
    \midrule
    S1 & 0.005638 & 0.045077 & 0.983533 \\
    S2 & 0.004462 & \textbf{0.039230} & 0.987040 \\
    S3 & 0.005426 & 0.045755 & 0.984630 \\
    S4 & \textbf{0.004381} & 0.039690 & \textbf{0.987258} \\
    \bottomrule
  \end{tabular}
\end{table}

\subsubsection{OGB-PCQM4Mv2}\label{sec:pcqm}

\begin{figure*}[!h]
  \centering
  \begin{subfigure}[t]{0.48\textwidth}
    \centering
    \includegraphics[width=\linewidth]{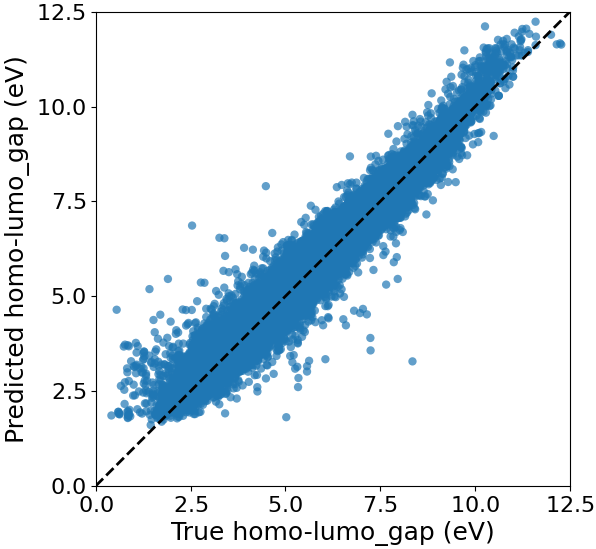}
    \caption{S1: GPS and Encoders disabled}
  \end{subfigure}\hfill
  \begin{subfigure}[t]{0.48\textwidth}
    \centering
    \includegraphics[width=0.99\linewidth]{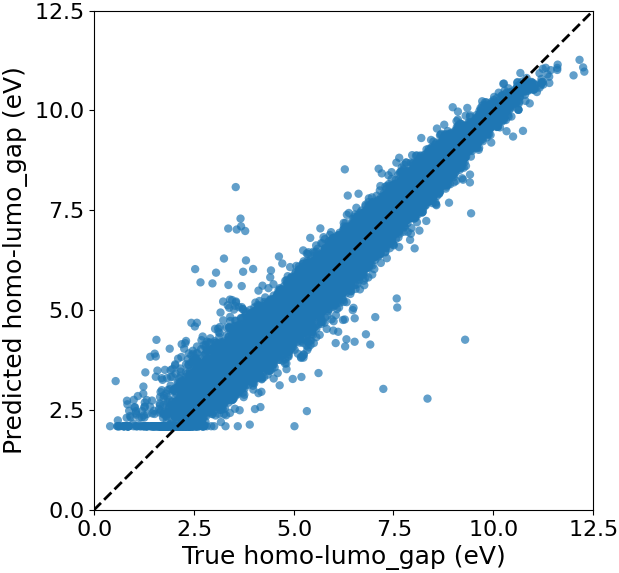}
    \caption{S2: GPS disabled, Encoders enabled}
  \end{subfigure}
  \vspace{0.6em}
  \begin{subfigure}[t]{0.48\textwidth}
    \centering
    \includegraphics[width=\linewidth]{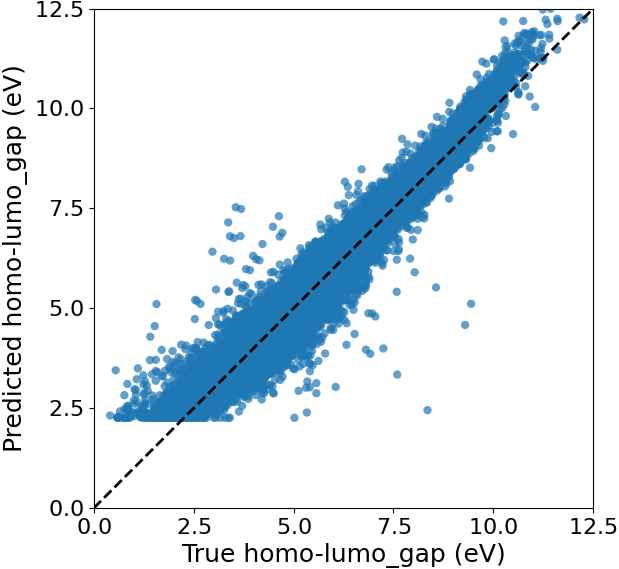}
    \caption{S3: GPS enabled, Encoders disabled}
  \end{subfigure}\hfill
  \begin{subfigure}[t]{0.48\textwidth}
    \centering
    \includegraphics[width=0.99\linewidth]{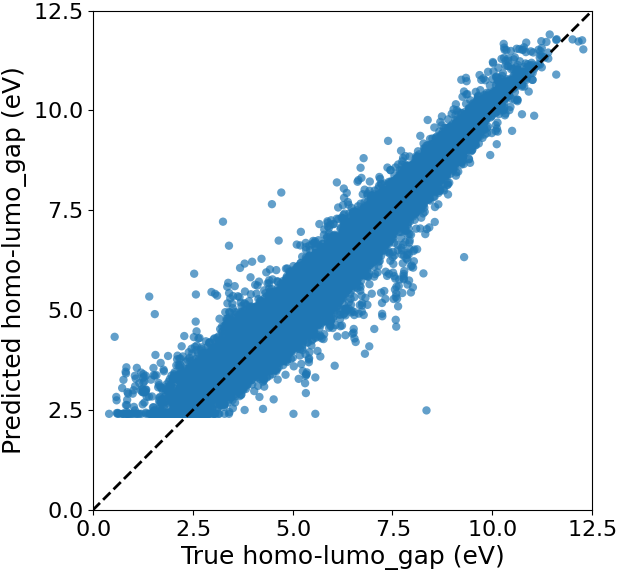}
    \caption{S4: GPS and Encoders enabled}
  \end{subfigure}
  \caption{OGB-PCQM4Mv2 parity plots (Predicted vs.\ True hom-lumo gap) for the best HPO trial per scheme}
  \label{fig:OGB-PCQM4Mv2-gap-scatter}
\end{figure*}

As shown in Table~\ref{tab:OGB-PCQM4Mv2-metrics}, the encoder-augmented PAINN without GPS (\textbf{S2}, $71.1$k parameters) delivers the best MSE/MAE and highest \(r\), while being smaller than the DimeNet baseline (\textbf{S1}, $95.1$k) and much smaller than the GPS-heavy model (\textbf{S4}, $130.2$k). Relative to \textbf{S1}, \textbf{S2} reduces MSE by \(\sim27\%\) with fewer parameters, and it outperforms \textbf{S4} by \(\sim18\%\) MSE despite a $45\%$ smaller footprint, highlighting strong parameter efficiency. The parity plots are shown in Figure~\ref{fig:OGB-PCQM4Mv2-gap-scatter}. \textbf{S2} shows the least widening of residuals in this regime, suggesting better calibration and smoothing of long-range interactions without the added complexity of GPS.

\begin{table}[!h]
  \centering
  \caption{OGB-PCQM4Mv2 best HPO configuration per scheme}
  \label{tab:OGB-PCQM4Mv2-schemes}
  \begin{tabular}{l
                  S[table-format=2.0]
                  S[table-format=2.0]
                  S[table-format=1.0]
                  S[table-format=2.0]
                  S[table-format=1.0]
                  S[table-format=1.0]
                  S[table-format=1.0]}
    \toprule
     \textbf{Scheme} & \textbf{MPNN} & \textbf{\#Conv} & \textbf{Hidden} & \textbf{EdgeEmb} & \textbf{GPS Heads} & \textbf{\#Parameters} \\
    \midrule
    S1 & \text{DimeNet} & 5 & 38 & 0 & 0 & 95051\\
    S2 & PAINN & 2 & 45 & 9 & 0 & 71084\\
    S3 & PAINN & 3 & 32 & 13 & 2 & 82280\\
    S4 & PAINN & 3 & 40 & 10 & 8 & 130203\\
    \bottomrule
  \end{tabular}
\end{table}

\begin{table}[!h]
  \centering
  \caption{OGB-PCQM4Mv2 test performance per scheme}
  \label{tab:OGB-PCQM4Mv2-metrics}
  \sisetup{round-mode=places,round-precision=6}
  \begin{tabular}{l
                  S[table-format=1.6]
                  S[table-format=1.6]
                  S[table-format=1.6]}
    \toprule
    \textbf{Scheme} & \textbf{MSE} $\downarrow$ & \textbf{MAE} $\downarrow$ & \textbf{Pearson $r$} $\uparrow$ \\
    \midrule
    S1 & 0.041413 & 0.141160 & 0.984880 \\
    S2 & {\bfseries 0.030317} & {\bfseries 0.124741} & {\bfseries 0.988888} \\
    S3 & 0.042750 & 0.146679 & 0.984439 \\
    S4 & 0.037168 & 0.135053 & 0.986146 \\
    \bottomrule
  \end{tabular}
\end{table}

\subsubsection{OGB-PPA}\label{sec:ppa}

Tables~\ref{tab:ogb-ppa-schemes} and~\ref{tab:ogb-ppa-accuracy} show that the best accuracy is obtained by a moderately deep GPS model (\textbf{S4}, 3 conv, 2 heads) at $369.8$k parameters, exceeding the strongest no-GPS baseline (\textbf{S1}, $86.2$k) by $1.77$ percentage points. However, the improvement comes at a $\sim 4\times$ parameter cost, and a deeper/wider no-GPS model (\textbf{S2}, $340.7$k) underperforms both \textbf{S1} and \textbf{S4}, indicating that capacity alone does not substitute for global information mixing. The shallow GPS model (\textbf{S3}, $65.1$k) underachieves, suggesting that GPS is effective only when paired with sufficient hidden/edge capacity and moderate depth. Overall, PPA benefits from GPS, but the accuracy-per-parameter trade-off still favors compact baselines in settings where parameter budget is constrained.

\begin{table}[!h]
  \centering
  \caption{OGB-PPA best HPO configuration per scheme}
  \label{tab:ogb-ppa-schemes}
  \begin{tabular}{l
                  S[table-format=2.0]
                  S[table-format=2.0]
                  S[table-format=1.0]
                  S[table-format=2.0]
                  S[table-format=1.0]
                  S[table-format=1.0]
                  S[table-format=1.0]}
    \toprule
    \textbf{Scheme} & \textbf{MPNN} & \textbf{\#Conv} & \textbf{Hidden} & \textbf{EdgeEmb} & \textbf{GPS Heads} & \textbf{\#Parameters} \\
    \midrule
    S1 & PNA   & 5 & 30 & 0  & 0 & 86151\\
    S2 & PNA  & 5 & 56 & 19  & 0 & 340728\\
    S3 & PNA  & 2 & 32 & 15  & 2 & 65128\\
    S4 & PNA  & 3  & 64 & 22  & 2 & 369804\\
    \bottomrule
  \end{tabular}
\end{table}

\begin{table}[!h]
  \centering
  \caption{OGB-PPA test accuracy per scheme}
  \label{tab:ogb-ppa-accuracy}
  \begin{tabular}{l S[table-format=2.2] S[table-format=2.2] S[table-format=2.2] S[table-format=2.2]}
    \toprule
     & \multicolumn{1}{c}{\textbf{S1}} & \multicolumn{1}{c}{\textbf{S2}} & \multicolumn{1}{c}{\textbf{S3}} & \multicolumn{1}{c}{\textbf{S4}} \\
    \midrule
    \textbf{Accuracy (\%)} $\uparrow$ & 65.24 & 63.75 & 53.24 & {\bfseries 67.01} \\
    \bottomrule
  \end{tabular}
\end{table}

\subsubsection{OGB-molPCBA}\label{sec:pcba}

For molPCBA, GPS with PNA improves both accuracy and parameter efficiency (Table~\ref{tab:OGB-molPCBA-schemes} and~\ref{tab:OGB-molPCBA-map}). The best configuration, \textbf{S3} ($3$ conv, $2$ heads), reaches mAP $=0.184$ with $215.8$k parameters—outperforming the deeper no-GPS PNA (\textbf{S2}, 252.1k; $+0.017$ absolute, $+10.2\%$ relative) and the larger GAT baseline (\textbf{S1}, 304.3k; $+0.029$, $+18.7\%$) while using fewer parameters. Increasing heads from $2$ to $4$ (\textbf{S4}, $217.0$k) yields no additional gain, consistent with diminishing returns from global capacity once moderate long-range mixing is available. These results indicate that, for molPCBA, moderate depth with limited GPS is sufficient to exploit cross-task structure, and further scaling primarily inflates model size without commensurate improvements.

\begin{table}[!h]
  \centering
  \caption{OGB-molPCBA best HPO configuration per scheme}
  \label{tab:OGB-molPCBA-schemes}
  \begin{tabular}{l
                  S[table-format=2.0]
                  S[table-format=2.0]
                  S[table-format=1.0]
                  S[table-format=2.0]
                  S[table-format=1.0]
                  S[table-format=1.0]
                  S[table-format=1.0]}
    \toprule
    \textbf{Scheme} & \textbf{MPNN} & \textbf{\#Conv} & \textbf{Hidden} & \textbf{EdgeEmb} & \textbf{GPS Heads} & \textbf{\#Parameters} \\
    \midrule
    S1 &  GAT   & 3 & 44 & 0 & 0 & 304270  \\
    S2 &  PNA   & 5 & 48 & 8 & 0 & 252106 \\
    S3 &  PNA  & 3 & 48 & 12 & 2 & 215829\\
    S4 &  PNA   & 3 & 48 & 12 & 4 & 216969\\
    \bottomrule
  \end{tabular}
\end{table}

\begin{table}[!h]
  \centering
  \caption{OGB-molPCBA test performance per scheme}
  \label{tab:OGB-molPCBA-map}
  \begin{tabular}{l S[table-format=1.3] S[table-format=1.3] S[table-format=1.3] S[table-format=1.3]}
    \toprule
     & \textbf{S1} & \textbf{S2} & \textbf{S3} & \textbf{S4} \\
    \midrule
    {\textbf{mAP} $\uparrow$} & 0.155 & 0.167 & {\bfseries 0.184} & 0.182 \\
    \bottomrule
  \end{tabular}
\end{table}

{\subsubsection{Practical improvements over the HydraGNN baseline}

To assess the practical impact of the proposed extensions, we compare all architectures against the HydraGNN baseline (Scheme S1), which corresponds to standard message passing without global attention or encoder-based feature augmentation. Across multiple datasets and tasks, we observe that attention-augmented and encoder-enhanced schemes (S2–S4) consistently match or outperform the HydraGNN baseline, with the largest gains occurring in datasets characterized by larger graph sizes or more complex structural dependencies. In particular, encoder-based augmentation (S2) and the combined scheme (S4) yield improved predictive accuracy on several benchmarks while maintaining comparable training stability, confirming that the proposed extensions can provide tangible performance benefits beyond the original HydraGNN architecture. These results indicate that incorporating attention mechanisms and feature encoders within a unified framework can enhance model expressivity and accuracy without sacrificing the scalability and robustness that motivate HydraGNN.

\subsubsection{Interpretation of HPO-selected hidden dimensions}

Across multiple datasets and architectures, the HPO process frequently selected relatively small hidden dimensions (often smaller than 64), indicating that wider representations did not consistently yield improved performance under the given training and data regimes. This behavior is consistent with prior observations that molecular graph datasets can saturate in performance at moderate model widths \cite{Hu2020Strategies, Gasteiger2020DimeNet}, particularly when combined with expressive message passing or attention mechanisms \cite{10.5555/3648699.3648742}.

\subsubsection{Results synthesis}
Rather than confirming or refuting a blanket limitation of message passing neural networks, our results show that their effectiveness depends strongly on how architectural choices, feature representations, and training protocols are combined. Across the evaluated datasets, well-tuned MPNNs augmented with domain and topological encoders frequently achieve performance comparable to, or better than, models relying solely on global attention. Global attention provides measurable benefits in specific regimes—particularly for tasks characterized by larger graphs or pronounced nonlocal dependencies—but is not uniformly superior. These findings highlight that claims regarding the necessity of global attention should be evaluated in a controlled setting and in relation to the underlying data characteristics.

}

\section{Discussion}\label{sec:disc}
{The goal of our work is not to claim that global attention mechanisms are universally necessary for atomistic graph learning, nor that message passing alone is fundamentally insufficient. Instead, our aim is to determine when global attention provides tangible benefits over well-tuned MPNN baselines, and under what data and architectural conditions those benefits justify their additional computational cost. By enabling controlled comparisons within a single HydraGNN framework, this study reframes commonly cited limitations of MPNNs as empirically testable hypotheses rather than assumptions. 

Widely cited limitations of MPNNs—particularly their difficulty in capturing long-range interactions—are often inferred from isolated case studies or comparisons across heterogeneous architectures, datasets, and training protocols. As a result, it remained unclear to what extent these limitations arise from the message passing paradigm itself versus confounding factors such as architectural choices, feature representations, or optimization settings. A central goal of our work was to systematically examine these claims through controlled experimentation. }

In contrast to prior work, our contribution is not to introduce yet another architecture, but to deliver the first unified framework for reproducible atomistic benchmarking that incorporates MPNNs, global attention, hybrid GPS-style models, and encoder-based feature augmentation under identical training, implementation, and hyperparameter optimization settings. {By implementing all combinations of message passing, global attention, and encoder-based feature augmentation within a single HydraGNN framework, we enabled direct and reproducible comparisons under identical training and hyperparameter optimization procedures. This unified setup allowed us to assess when standard MPNNs are sufficient and when additional mechanisms provided measurable benefits, thereby transforming qualitative assumptions about long-range modeling into empirically testable hypotheses}. This design enables us to isolate the effect of each modeling component—local message passing, global attention, and encoders—across diverse datasets, providing principled guidance on when attention mechanisms are truly beneficial. To our knowledge, no existing study has provided such a controlled evaluation.

Across datasets, our experiments clarify when encoder augmentations and global attention (GPS) provide measurable benefit, and at what computational cost. On small to mid–scale regression tasks (ZINC, QM9, TMQM), encoder–augmented PAINN without GPS is consistently competitive or best: on ZINC and QM9, the encoder–only setting (S2) yields the lowest errors while remaining compact (Tables~\ref{tab:zinc-hpo}–\ref{tab:zinc-metrics}, \ref{tab:qm9-schemes}–\ref{tab:qm9-metrics}), and on TMQM a shallow PAINN with encodings (S2) again minimizes MSE/MAE despite being smaller than GPS variants (Tables~\ref{tab:tmqm-schemes}–\ref{tab:tmqm-metrics}). Parity plots (Figs.~\ref{fig:zinc-parity-4up}, \ref{fig:qm9-freeenergy-scatter}, \ref{fig:tmqm-dispersion-scatter}) show correlations near unity across schemes, with residual differences concentrated in a few high-magnitude outliers; in this regime, richer node/edge encodings deliver the dominant gains, and additional global information offers little benefit at similar budgets. In contrast, tasks with weaker local features or pronounced long-range dependence (OGB-PPA, OGB-PCBA) benefit moderately on use of GPS when paired with adequate width and edge capacity: on PPA, a moderately deep PNA+GPS model (S4; three convolutional layers, two GPS heads) improves top-1 accuracy over the strongest no-GPS baseline at higher parameter cost (Tables~\ref{tab:ogb-ppa-schemes}–\ref{tab:ogb-ppa-accuracy}), while the shallower GPS model (S3) underperforms; on PCBA, PNA+GPS with two heads (S3) attains the highest mAP with fewer parameters than both the deeper no-GPS PNA (S2) and the larger GAT baseline (S1), and increasing heads to four (S4) yields diminishing returns (Tables~\ref{tab:OGB-molPCBA-schemes}–\ref{tab:OGB-molPCBA-map}). Collectively, these results indicate that moderate global information (two heads) and moderate depth (three message-passing layers) are sufficient to capture nonlocal effects in large, atomistic graphs, whereas encoder quality is the primary driver on chemically local regressions. {The frequent observation that encoder-only models (S2) outperform GPS-only or fused variants (S3–S4) is not a limitation of the framework, but a central empirical finding enabled by it, showing that architectural fusion does not guarantee improved performance without complementary feature representations.}

A second aspect is {parameter efficiency}. Across datasets, the best or near-best schemes typically fall in the tens to a few hundreds of thousands of parameters. For instance, on OGB-PCQM4Mv2, the encoder-augmented PAINN without GPS (S2; $71$k parameters) achieves the lowest MSE/MAE and the highest correlation, outperforming both a deeper DimeNet baseline (S1; $95$k) and a larger GPS model (S4; $130$k) (Tables~\ref{tab:OGB-PCQM4Mv2-schemes}–\ref{tab:OGB-PCQM4Mv2-metrics}; Fig.~\ref{fig:OGB-PCQM4Mv2-gap-scatter}). On PPA, the accuracy gain of S4 over S1 is modest relative to the \(>4\times\) parameter increase; thus the baseline without GPS remains useful when the parameter budget is constrained. On NIAID, GPS yields small but consistent improvements in MSE and correlation (S4), whereas the lowest MAE is achieved without GPS (S2), underscoring that the best configuration can depend on the target metric (Tables~\ref{tab:niaid-mof-schemes}–\ref{tab:niaid-mof-metrics}; Fig.~\ref{fig:niaid-mof-partialcharge-scatter}).

GraphGPS~\citep{rampasek2022gps} popularized the hybrid recipe of local MPNN layers with GPS blocks and rich positional/structural encodings. The reference configurations reported for ZINC, PPA, PCBA, and PCQM4Mv2 are strong but operate at substantially higher capacity and depth than our study; for example, on PCQM4Mv2 the \emph{gps-medium} setting employs 10 GPS layers with a hidden size in the mid–hundreds and a multi–million parameter budget (e.g., $\sim 9.5$M parameters), and even “small’’ variants remain in the multi–million regime. These designs achieve excellent absolute scores but can be burdensome to reproduce exactly, owing to large memory footprints and sensitivity to hyperparameters and training infrastructure. Our experiments intentionally adopt small models with transparent toggles (encoders on/off; GPS on/off; heads and depth) and report parameter counts alongside metrics. This capacity-aware protocol exposes when GPS truly helps (PPA/PCBA) and when encoders alone suffice (ZINC/QM9/PCQM), and it facilitates fairer, more reproducible comparisons than matching against highly tuned, much larger systems.

For datasets with established SOTA baselines (ZINC, PCQM4v2, molPCBA etc.), we refrain from emphasizing raw numerical gaps to specific leader-boards and instead propose a \emph{normalized, budget-controlled} evaluation: every result is paired with parameter counts and layer budgets (Tables above), and parity plots visualize the error structure beyond single-number summaries. For datasets without widely used SOTA leader-boards (TMQM, NIAID etc.), we provide compact, reproducible references together with parity plots, establishing a consistent yardstick for future work. Across tasks, three practical takeaways emerge: (i) encoder modules account for most of the gains on chemically local regressions; (ii) moderate GPS (two heads) improves classification on large, atomistic graphs when paired with adequate width and edge embeddings; and (iii) scaling GPS depth/heads beyond this point yields diminishing returns relative to parameter growth. Overall, exposing both architectural toggles and parameter counts offers a clearer and more reproducible basis for assessing progress than relying on large, in-replicable configurations.

{
Table~\ref{tab:when_gps_helps} summarizes our findings across dataset characteristics.
\begin{table}[h]
\centering
\caption{Decision framework: when to use global attention based on dataset characteristics}
\label{tab:when_gps_helps}
\small
\begin{tabular}{|l|l|l|p{5.5cm}|}
\hline
\textbf{Graph Size} & \textbf{Task Type} & \textbf{Config} & \textbf{Examples} \\
\hline
Small & Regression & S2 & Local chemistry dominates. Encoders \\
($<$50 nodes) & & (MPNN+Enc) & capture atomic/bond features. \\
& & & \textit{QM9, ZINC, PCQM4Mv2} \\
\hline
Medium & Regression & S2 & Encoders still effective for local \\
(50--150) & & (MPNN+Enc) & properties. \textit{TMQM (120 nodes)} \\
\hline
Large & Classification & S3/S4 & Global context aids discrimination \\
($>$150 nodes) & & (GPS 2 heads) & across distant substructures. \\
& & & \textit{OGB-PPA (243 nodes)} \\
\hline
Small-Med & Multi-label & S3 & Multi-label learning benefits from \\
& Classification & (GPS) & global information. \\
& & & \textit{OGB-PCBA (26 nodes)} \\
\hline
Large & Node-level & S4 & Long-range charge/field effects. \\
& Regression & (GPS+Enc) & Encoders + GPS both help. \\
& & & \textit{NIAID (200 nodes)}. S2 is also competitive in terms of MAE for this use case. \\
\hline
\end{tabular}
\end{table}
We observe the following patterns: (1) graph size alone doesn't determine GPS utility—task type matters, (2) for regression on local molecular properties, encoders suffice regardless of moderate size increases, (3) classification tasks benefit more from GPS, especially on large graphs, and (4) when GPS helps, 2 heads with 3 conv layers is enough; more capacity yields diminishing returns. These guidelines derive from seven datasets and thresholds are approximate; for novel chemical systems, it will be crucial to validate empirically.
}

\section{Conclusions and Future Work}\label{sec:conc}

\noindent \textbf{Conclusions.} We presented a capacity–controlled, reproducible evaluation of local (MPNN), hybrid, and fused local–global graph architectures within a unified HydraGNN pipeline, isolating the effects of encoders and global attention (GPS) across regression and classification benchmarks. Three findings emerge. \textit{First}, domain encoders are the dominant lever on chemically local regressions: on ZINC, QM9, TMQM, and PCQM, encoder-augmented PAINN without GPS consistently matches or exceeds alternatives while remaining compact, and parity plots indicate that remaining errors are concentrated in a few high-magnitude tails rather than systematic miscalibration. \textit{Second}, GPS complements rather than replaces message passing: on OGB-PPA and OGB-PCBA, moderate global information mixing—two GPS heads paired with three message-passing layers and adequate hidden/edge capacity—yields the most reliable gains, whereas adding more heads or depth exhibits diminishing returns and under-sized GPS models underperform. \textit{Third}, parameter efficiency matters: strong results are obtained with tens to a few hundreds of thousands of parameters, often outperforming deeper baselines and approaching or surpassing much larger GPS variants; reporting metrics alongside parameter counts exposes the true accuracy–capacity trade-off and improves comparability. Practically, we recommend encoder-augmented MPNNs as the default for 
small-to-medium molecular regression tasks (avg. nodes $<50$), enabling GPS 
when (i) graphs are large (avg. nodes $> 150$), (ii) tasks involve classification, 
or (iii) targets depend on explicit long-range interactions spanning many bonds 
or significant spatial distances. When GPS is warranted, start with small budgets 
(two heads, three layers) before scaling width or heads. For intermediate cases 
(50-150 nodes, mixed local/nonlocal effects), empirical comparison between S2 
and S4 configurations is recommended. Beyond established leaderboards, we provide compact, transparent baselines and diagnostic plots for datasets lacking widely used references, offering a reproducible yardstick for future work and a clearer path to principled scaling under fixed compute budgets.\\

\noindent \textbf{Future work.} The present encoders omit explicit geometric features such as positional, distance and angular information; incorporating these is likely to improve geometry-sensitive tasks. Also, our current transformer model does not preserve equivariance which can be necessary for multiple physics-informed downstream tasks. 
Further, we did not study large-scale pretraining or sparse/linearized attention variants that could reduce the quadratic overhead. Extending the evaluation to larger bio-molecular and crystalline systems, adding uncertainty quantification, and exploring equivariant attention for vector/tensor targets are natural next steps. We expect the framework and results to serve as a reference point for method development and for principled decisions about when global attention is useful in atomistic graph learning.

\subsection{Data Availability Statement}

All datasets and all scripts used in this study are available on the GitHub repository:
\href{https://github.com/ORNL/HydraGNN}{HydraGNN}

\section*{Authors' Contributions}
A.~C. developed the GPS implementation in HydraGNN, ran the experiments described in this paper, and drafted the manuscript. M.~L.~P. mentored A.~C. while performing the research and contributed to the writing of the manuscript. 

\section*{Acknowledgement}
This research is sponsored by the Artificial Intelligence Initiative as part of the Laboratory Directed Research and Development (LDRD) Program of Oak Ridge National Laboratory, managed by UT-Battelle, LLC, for the US Department of Energy under contract DE-AC05-00OR22725.
This work used resources of the Oak Ridge Leadership Computing Facility, which is supported by the Office of Science of the U.S. Department of Energy under Contract No. DE-AC05-00OR22725, under ASCR Leadership Computing Challenge (ALCC) award LRN070. This work also used resources of the National Energy Research Scientific Computing Center (NERSC), a Department of Energy User Facility using NERSC award ScienceAtScale@NERSC.

\bibliography{sn-bibliography}

\end{document}